\documentclass[review]{elsarticle}

\usepackage{lineno,hyperref}
\usepackage{graphicx}

\usepackage{amsmath,amssymb}
\usepackage{mathrsfs}
\usepackage{color}
\usepackage{graphics}
\usepackage{multirow}
\usepackage{caption}
\usepackage{subcaption}
\usepackage{amssymb}
\usepackage{array}
\usepackage{amsmath,bm}
\usepackage{colortbl}
\usepackage{dsfont}
\usepackage{diagbox}
\usepackage{rotating}
\usepackage{booktabs}
\usepackage{overpic}
\usepackage{contour}
\usepackage{bbding}
\usepackage[marginal]{footmisc}
\usepackage{pifont}
\usepackage{pifont}
\usepackage{booktabs}
\usepackage{makecell}
\usepackage{verbatim}
\usepackage{bbding}
\usepackage{url}

\usepackage{algorithm}
\usepackage{algorithmic}

\definecolor{mygray}{gray}{.92}
\definecolor{mypink}{RGB}{251, 81, 169}

\graphicspath{{./Imgs/}}
\DeclareGraphicsExtensions{.jpg,.pdf,.png}

\newcommand{\eg}[1]{\textit{e.g.,}}
\newcommand{\ie}[1]{\textit{i.e.,}}

\newcommand{\fig}[1]{Fig.~\ref{#1}}

\newcommand{\revise}[1]{\textcolor{black}{#1}}
\modulolinenumbers[5]

\journal{Journal of \LaTeX\ Templates}

\begin{document}

\begin{frontmatter}

\title{Fully Context-Aware Image Inpainting with a Learned Semantic Pyramid}
% \tnotetext[mytitlenote]{Fully documented templates are available in the elsarticle package on \href{http://www.ctan.org/tex-archive/macros/latex/contrib/elsarticle}{CTAN}.}

\author{Wendong Zhang}
\author{Yunbo Wang\corref{mycorrespondingauthor}}\cortext[mycorrespondingauthor]{Corresponding author}\ead{yunbow@sjtu.edu.cn}
\author{Bingbing Ni}
\author{Xiaokang Yang}
\address{MoE Key Lab of Artificial Intelligence, AI Institute, Shanghai Jiao Tong University, Shanghai, China.}

\begin{abstract}
Restoring reasonable and realistic content for arbitrary missing regions in images is an important yet challenging task. 
Although recent image inpainting models have made significant progress in generating vivid visual details, they can still lead to texture blurring or structural distortions due to contextual ambiguity when dealing with more complex scenes.
To address this issue, we propose the Semantic Pyramid Network (SPN) motivated by the idea that learning multi-scale semantic priors from specific pretext tasks can greatly benefit the recovery of locally missing content in images.
SPN consists of two components. First, it distills semantic priors from a pretext model into a multi-scale feature pyramid, achieving a consistent understanding of the global context and local structures.
Within the prior learner, we present an optional module for variational inference to realize probabilistic image inpainting driven by various learned priors.
The second component of SPN is a fully context-aware image generator, which adaptively and progressively refines low-level visual representations at multiple scales with the (stochastic) prior pyramid. 
We train the prior learner and the image generator as a unified model without any post-processing.
Our approach achieves the state of the art on multiple datasets, including Places2, Paris StreetView, CelebA, and CelebA-HQ, under both deterministic and probabilistic inpainting setups.

\end{abstract}


\end{frontmatter}

%\linenumbers

%%%%%%%%%%%%%%%%%%%%%%%%%%%%%%%%%%%%%%%%%%%%%%%%%%%%%%%%%%%%%%%%%%%%
% introduction
%%%%%%%%%%%%%%%%%%%%%%%%%%%%%%%%%%%%%%%%%%%%%%%%%%%%%%%%%%%%%%%%%%%%
\section{Introduction}
Inpainting is an important image processing method, which is widely used in many applications including image restoration~\cite{DBLP:journals/pr/DingRR18}, object removal~\cite{criminisi2003object}, and photo editing~\cite{barnes2009patchmatch}. %, and image compression~\cite{DBLP:journals/tcsv/LiuSWLZ07,DBLP:journals/tcsv/QinZCDZ19}.criminisi2004region
Specifically, given the damaged image and the corresponding mask, image inpainting aims to synthesize the plausible visual content for the missing area, and make the recovered content consistent with the remaining area.
In general, the key challenge in generating meaningful and clear visual details lies in how to properly harness the context ambiguity between low-level structures and high-level semantics.

\begin{figure*}[t]
\begin{center}
\includegraphics[width=1.0\textwidth]{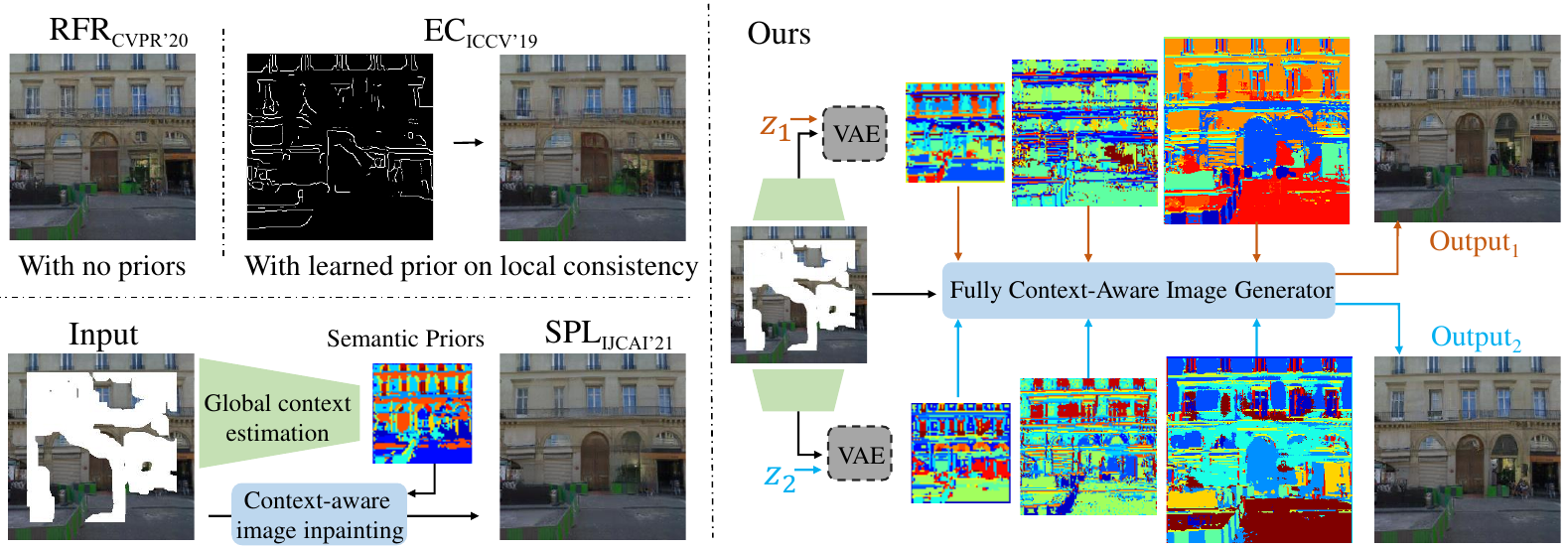}
\end{center}
\vspace{-2mm}
\caption{
\textbf{Model comparison. Top Left:} Existing image inpainting models, \eg, RFR~\cite{li2020recurrent} and EC~\cite{nazeri2019edgeconnect}, mainly focus on local textures and often suffer from image distortion due to semantic ambiguity.
\textbf{Bottom left:} Our previous work, \textit{semantic prior learner} (SPL)~\cite{DBLP:conf/ijcai/ZhangZTWCNWY21} introduces deterministic semantic guidance at a single scale.
\textbf{Right:} SPN learns a pyramid of semantic priors with an optional variational module. It not only improves contextual reasoning at multiple visual scales but also enables the model to handle probabilistic image inpainting.
}
\vspace{-2mm}
\label{motivation}
\end{figure*}

Existing approaches focus more on the consistency of local textures.
Earlier work~\cite{barnes2009patchmatch} usually %, such as diffusion-based or patch-based approaches~\cite{barnes2009patchmatch,DBLP:conf/iccv/LevinZW03,DBLP:journals/tip/RuzicP15}, ,DBLP:journals/tip/RuzicP15,DBLP:conf/iccv/LevinZW03
exploit low-level similarity to perform texture transformations from valid image regions to missing regions. These methods work well in stationary backgrounds but usually fail to handle complex structures due to a lack of global semantic understanding. 
In recent years, the development of deep learning has stimulated the emergence of many image inpainting methods in forms of neural networks~\cite{yu2020region,li2020recurrent,nazeri2019edgeconnect,DBLP:conf/eccv/LiuJSHY20}. 
%,xie2019image
These methods %are usually based on generative adversarial network (GAN)~\cite{goodfellow2014generative} or variational auto-encoder (VAE)~\cite{DBLP:journals/corr/KingmaW13} and 
try to solve the image inpainting problem by using the image-to-image translation framework. 
In particular, some literature~\cite{DBLP:conf/eccv/LiuJSHY20,nazeri2019edgeconnect} have introduced different structural guidance to specifically supervise the completion of missing areas. %, including color-smooth images~\cite{DBLP:conf/eccv/LiuJSHY20} and edge maps~\cite{nazeri2019edgeconnect}.
Although the learning-based methods remarkably improve the quality of the generated images, as shown in \fig{motivation}, they still show distorted objects and blur textures, especially for complex scenes. 
An important reason is that these methods focus more on pixel-level image representations, but do not explicitly consider semantic consistency of context representations across multiple visual levels, and thus fail to establish a global understanding of the corrupted images.

In contrast to the above literature, we propose to acquire multi-scale semantic priors from specific pretext tasks and leverage them in image inpainting.
Motivated by this idea, we present a fully context-aware image inpainting model named Semantic Pyramid Network (SPN), which contains two main components. 
The first one is the semantic prior learner, which distills a pyramid of visual representations from different levels of a pre-trained multi-label classification model\footnote{Various pretext tasks can be used to provide semantic supervisions, whose effect is compared in the experiments. In the rest of the paper, we adopt multi-label classification as a typical pretext task without loss of generality.}. 
By optimizing the semantic learner in the entire image reconstruction pipeline, we enable the model to adaptively transfer the common knowledge from the pretext task.
In this process, the high-level visual priors contain more discriminative representations and thus can effectively reduce the ambiguity of global semantics (see \fig{motivation}), while the low-level visual priors provide further guidance for restoring the local textures.
Based on the learned \textit{semantic prior pyramid}, we present the second component of SPN, that is, the fully context-aware image generator.
Instead of directly concatenating the prior information with image features extracted from pixels, we use the spatially-adaptive normalization~\cite{park2019semantic} to gradually integrate them into different levels of the generator for global context reasoning and local image completion. 
In this way, SPN allows for more flexible interactions between the learned priors and image features, avoiding potential domain conflicts caused by the use of pretext models from another domain. 

This paper presents two new technical contributions that effectively complement our previous work at IJCAI'21~\cite{DBLP:conf/ijcai/ZhangZTWCNWY21}: First, as mentioned above, we extend the prior learner to distill semantic presentations at different scales from the pretext model. Accordingly, we further enable the image generator to incorporate the prior pyramid. By this means, we improve the consistency of global context reasoning and local image completion.
Second, we extend our model to probabilistic image inpainting by integrating a plug-and-play variational inference module in the prior learner.
In addition to knowledge distillation loss, the prior learner is also optimized to fit the distribution of missing content in a probabilistic way. 
The main idea is that according to the context of the remaining part of the image, we should consider various semantic possibilities of the missing part.
Notably, the multi-label classification may convey diversified semantic relations among multiple objects, which motivates us to use a probabilistic prior learner to extract different relational priors according to certain conditional information.
As a result, the stochastic prior pyramid allows the image generator to fill in the missing content in diverse and reasonable ways.

In summary, the key insight of this paper is to show that learning semantic priors from specific pretext tasks can greatly benefit image inpainting. We further strengthen the modeling ability of the learned priors in SPN from the following aspects:
% \begin{itemize}%[leftmargin=*]
% \item Multi-scale semantic priors are learned and organized in a feature pyramid to achieve consistent understanding of the global context and local structures. 
% \item The prior learner is allowed to be trained in a probabilistic manner. It is inspired by the idea that in stochastic image inpainting, the recovery of potential missing content should be driven by the diversity of semantic priors.
% \end{itemize}
1) Multi-scale semantic priors are learned and organized in a feature pyramid to achieve consistent understanding of the global context and local structures. 2) The prior learner is allowed to be trained in a probabilistic manner. It is inspired by the idea that in stochastic image inpainting, the potential missing content should be driven by the diversity of semantic priors.

In the rest of this paper, we first review the related work on image inpainting and knowledge distillation in Section~\ref{related_work}, and introduce the details of our model SPN as well as its deterministic and probabilistic learning schemes in Section~\ref{approach}.
In Section~\ref{experiments}, we perform in-depth empirical analyses of the proposed SPN and compare it with the state of the art on four datasets. 
More importantly, we extend the experiments in our preliminary work by qualitatively and quantitatively evaluating the model under stochastic image inpainting setups. Further, we compare the effectiveness of learning the semantic prior pyramid using different types of pretext models.
In Section~\ref{conclusion}, we present the conclusions and the prospect of future research.

%%%%%%%%%%%%%%%%%%%%%%%%%%%%%%%%%%%%%%%%%%%%%%%%%%%%%%%%%%%%%%%%%%%%
% Related work
%%%%%%%%%%%%%%%%%%%%%%%%%%%%%%%%%%%%%%%%%%%%%%%%%%%%%%%%%%%%%%%%%%%%
\section{Related Work}
\label{related_work}
\subsection{Rule-Based Image Inpainting}

Early work for image inpainting mainly focuses on capturing pixel relationships using various low-level image features and designing specific rules to perform texture reasoning. %~\cite{DBLP:journals/tip/RuzicP15}.
These methods can be roughly divided into two groups.
The first one includes the patch-based methods, which assume that missing image regions can be filled with similar patches searched from valid image regions or external database~\cite{criminisi2003object,barnes2009patchmatch,DBLP:journals/tog/SunYJS05}. 
%,DBLP:conf/cvpr/BaekCK16,DBLP:journals/tip/DingRR19,DBLP:journals/tog/DroriCY03,DBLP:journals/npl/ElHarroussAAA20,criminisi2004region,,DBLP:conf/iccv/LevinZW03
%Following this line, many approaches were proposed to find better and faster patch matching algorithms~\cite{barnes2009patchmatch,DBLP:journals/tog/SunYJS05}. 
%,DBLP:conf/cvpr/BaekCK16,DBLP:journals/tip/DingRR19
However, these algorithms are often computationally expensive, and similar image patches are difficult to access for complex scenes.
The second group includes diffusion-based methods that progressively propagate visual content from boundary regions into missing regions~\cite{DBLP:conf/iccv/BallesterCVBS01}. %,DBLP:conf/cvpr/BertalmioBS01,ballester2001filling,DBLP:conf/iccv/LevinZW03,
These methods are usually based on specific boundary conditions and use partial differential equations to guide the pixel propagation.
%
%Also, there is existing approach that jointly use patch-based and diffusion-based techniques~\cite{DBLP:journals/tip/Bertalmi03}. 
% ,DBLP:journals/pr/DingRR18
Although these methods can produce vivid visual details for the plain background and repeated textures, they often fail to generate realistic image content for complex scenes because only low-level visual structures are modeled without any semantic understanding.
% Rule-based methods can produce vivid visual details for the restoration of a plain background and repeated textures.
% %
% However, they often fail to generate realistic image content for complex scenes because only low-level visual structures are modeled without any semantic understanding.

\subsection{Learning-Based Image Inpainting}

\subsubsection{Deep networks for deterministic image inpainting}
In recent years, many deep learning models based on the architectures of conditional GAN have been proposed for deterministic image inpainting~\cite{xiong2019foreground,yu2018generative,DBLP:journals/pr/DingRR18,DBLP:journals/pr/ZhangWSYLKLZLYS22,DBLP:conf/cvpr/0002OC21}. %DBLP:conf/eccv/YanLLZS18,
These models also greatly benefit from the adversarial training scheme as well as further designs.
Pathak \textit{et al.}~\cite{pathak2016context} first exploited the auto-encoder framework to perform global context encoding and image decoding. %, which introduces a new channel-wise fully connected layer to learn the global context.
%
%On top of this, other learning-based methods have achieved promising improvements in restoring particular images such as human faces and cars~\cite{DBLP:conf/cvpr/Yeh0LSHD17, DBLP:journals/pr/ZhangWSYLKLZLYS22}. 
%
Iizuka \textit{et al.}~\cite{iizuka2017globally} proposed to use multiple discriminators based on stacked dilated convolutional layers to enhance the consistency of global and local visual context. %,DBLP:journals/pami/ChenPKMY18 ~\cite{DBLP:journals/corr/YuK15}
Other architectures are also explored in recent work %, including U-Net~\cite{DBLP:conf/eccv/YanLLZS18}, %the multi-column fusion module~\cite{DBLP:conf/nips/WangTQSJ18}, 
such as the multi-scale learning modules~\cite{li2020recurrent,QIN2022108547}. %to enlarge the global perception field of the image generator.
% wang2019musical,DBLP:journals/corr/abs-2104-01431,
Unlike the existing work, our model specifically leverages the multi-scale information as the semantic prior pyramid, which is learned in a multi-level knowledge distillation module.

More recent studies take advantage of the attention module in capturing long-range contextual relations~\cite{yu2018generative,DBLP:conf/cvpr/ZengFCG19,DBLP:journals/pr/WangMLZZ20}. %,DBLP:conf/eccv/SongYLLHLK18, liu2019coherent,sagong2019pepsi, ,xie2019image
For instance, Deepfill~\cite{yu2018generative} follows a coarse-to-fine generation framework with %a context attention module that relies on 
patch-wise aggregation to enhance the restoration of high-frequency information.
Besides, some methods propose various network architectures to eliminate the undesired influence from invalid pixels in context feature extraction~\cite{yu2019free,yu2020region}. %, such as partial convolutions~\cite{liu2018image}, gated convolutions~\cite{yu2019free}, and region normalization blocks~\cite{yu2020region}. %and dynamic selection blocks~\cite{DBLP:journals/tip/WangZZ21}. 
Although these approaches perform generally better than rule-based approaches at generating realistic images, they still struggle with structural distortion in complex scenes. This is because they cannot deal well with the ambiguous context of the missing area due to the limitation of the deterministic training paradigm.

\vspace{5pt}
\subsubsection{Probabilistic image inpainting models}
To cope with the multi-modal data distribution between the corrupted input images and the expected output images, different probabilistic models have been proposed, which can model the probability of the potential content in the missing area and produce diverse results.
PICNet~\cite{zheng2019pluralistic} exploits the framework of VAE in two different network branches to learn the conditional distribution within the missing areas.
%DBLP:conf/cvpr/ZhengCC19
UCTGAN~\cite{DBLP:conf/cvpr/ZhaoMLWZCXL20} also uses a similar architecture with the manifold projection module. 
Also, there are GAN-based methods~\cite{DBLP:conf/cvpr/LiuWHSH021,DBLP:journals/pr/ZengGZ21} and auto-regressive methods~\cite{DBLP:conf/cvpr/Peng0XL21} that explicitly consider the diversity of generated images. %DBLP:conf/cvpr/LahiriJAMB20,
Unlike all of the above methods, our model is particularly designed to harness the multi-modal distribution by leveraging a plug-and-play variational inference module in the learning process of semantic priors, instead of directly using it in the image generator.

% Instead of directly modeling the context distribution in low-level space, our semantic learner aim to consider various semantic possibilities for the missing regions with the help of the pre-trained pretext models. Combined with the proposed context-aware image generator, our model can produce more realistic and diverse results. 

\vspace{5pt}
\subsubsection{Visual priors in image inpainting}

Some state-of-the-art approaches also use explicit visual priors as the learning supervision to improve the quality of image inpainting. % especially for the clarity of the generated object structures. 
Nazeri \textit{et al.}~\cite{nazeri2019edgeconnect} and Xiong \textit{et al.}~\cite{xiong2019foreground} presented two-stage frameworks that first learn to generate the images of object edges under external supervisions, and then use them to guide texture restoration. 
%Xu \textit{et al.}~\cite{DBLP:journals/tcsv/XuLX21} further used a edge detector to enhance the edge completion results.
%
Some methods also use color-smooth images~\cite{ren2019structureflow,DBLP:conf/eccv/LiuJSHY20, guo2021image} or semantic maps~\cite{DBLP:conf/eccv/LiaoXWLS20} to provide more dense, structural, and meaningful supervisions. %DBLP:conf/bmvc/SongYSWHK18,
However, these methods either suffer from the accumulated errors through the two processing stages, or require heavy human annotations, which is difficult to scale to real-world applications.
In contrast, first, the semantic priors in our model are jointly learned with the image generator, % to distill useful prior knowledge adaptively, 
which greatly reduce error accumulation; Second, the priors are learned from the pyramid of feature maps of a pretext model, which are more accessible. % than the heavily annotated semantic maps. 
The concept of knowledge distillation was first proposed in the work from Hinton \textit{et al.}~\cite{DBLP:journals/corr/HintonVD15} to transfer pre-learned knowledge from a large teacher network to a smaller student network.
Although Suin \textit{et al.}~\cite{DBLP:conf/iccv/SuinP021} also used a joint-training autoencoder as distillation target, it lacks pre-learned semantic knowledge which are important for structural guidance.
Here, we adapt the knowledge distillation framework to high-level semantic priors and low-level texture priors simultaneously from the pretext model, both of which are important for context-aware image inpainting.

\section{Approach}
\label{approach}

\begin{figure*}[t]
    \centering
    \includegraphics[width=1.0\linewidth]{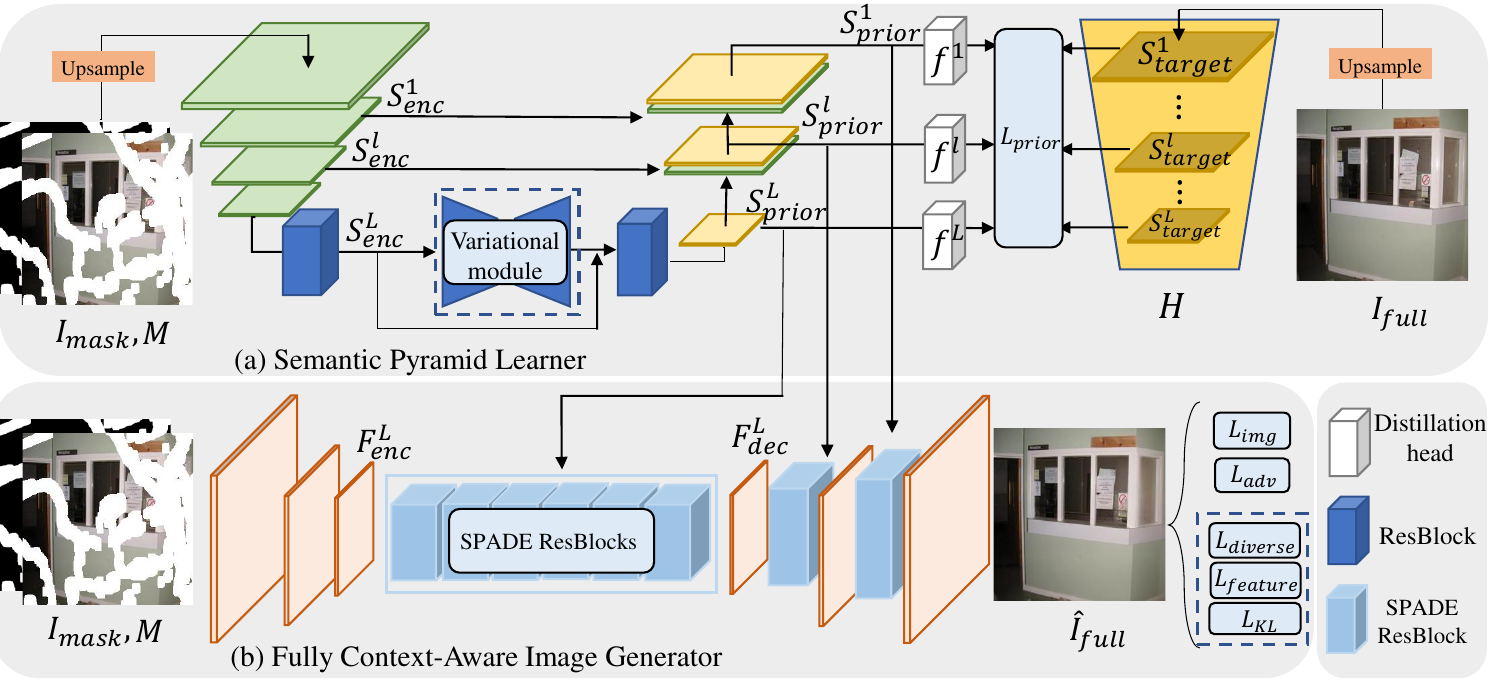}
    \vspace{-3mm}
    \caption{\textbf{Overview of SPN. (a)} The prior learner distills multi-scale semantic priors $\boldsymbol{S}^{l}_\text{prior}$ from $\boldsymbol{S}^{l}_\text{target}$ of the pretext multi-label classification model~\protect\cite{ben2020asymmetric} denoted by $H$. The semantic learner can generate diverse priors through an optional variational inference module. % shown in the dashed box. 
    \textbf{(b)} The image generator incorporates the learned priors in different levels of the network through \textit{spatially-adaptive normalization} modules (SPADE)~\protect\cite{park2019semantic}. The objective functions in the dashed box are optional and correspond to the probabilistic image inpainting setup.
    }
    % \vspace{-1mm}
    \label{Overview}
\end{figure*}

In this section, we introduce the details of the Semantic Pyramid Network (SPN) which aims to learn and transfer semantic priors from a particular pretext model to improve image inpainting. 
We emphasize that, in SPN, the semantic priors have the following properties: First, they should be learned and presented in a multi-scale feature pyramid to improve conditional image generation by reducing inconsistency between the global context and local textures.
Second, they should be endowed with a probabilistic training procedure, which enables SPN to generate diverse inpainting results by extracting various semantic priors from an individual damaged image.
The overall architecture of SPN is shown in \fig{Overview}. It mainly consists of two network components:
% \begin{itemize} %[leftmargin=*]
%     \item \textbf{(Stochastic) semantic prior learner:} It extracts multi-scale visual priors into a feature pyramid from a pretext model. It also exploits an optional, variational inference module to handle the stochasticity of image inpainting.
%     \item \textbf{Fully context-aware image generator:} It performs the affine transformation mechanism based on SPADE~\cite{park2019semantic} to progressively and adaptively integrate the learned semantic priors in the process of image restoration.
% \end{itemize}
1) \textbf{(Stochastic) semantic prior learner:} It extracts multi-scale visual priors into a feature pyramid from a pretext model. It also exploits an optional, variational inference module to handle the stochasticity of image inpainting. 2) \textbf{Fully context-aware image generator:} It performs the affine transformation mechanism based on SPADE~\cite{park2019semantic} to progressively and adaptively integrate the learned semantic priors in the process of image restoration.
% \end{itemize}

% In Section~\ref{semantic_pyramid}, we first discuss the respective forms of the deterministic and stochastic semantic prior learner.
% %
% Then, in Section~\ref{context-aware_generator}, we detail the image generator that incorporates multi-scale visual priors. 
% %
% Finally, in Section~\ref{loss_function}, we describe the overall training objectives of SPN.
% %
% It is worth noting that, in each of the above components, we make substantial improvements over our previous work at IJCAI'21~\cite{DBLP:conf/ijcai/ZhangZTWCNWY21}.

% \yb{emphasize the new contribution at the end of each subsection.}

\subsection{Semantic Prior Learner}
\label{semantic_pyramid}

\subsubsection{Learning deterministic prior pyramid}
As shown in~\fig{Overview}(a), the semantic prior learner follows the U-Net architecture. %~\cite{DBLP:conf/miccai/RonnebergerFB15}. 
It distills multi-scale visual priors from a particular pretext model given a full image $\boldsymbol{I}_\text{full}\!\in\!\mathbb{R}^{3\times h\times w}$, a damaged image $\boldsymbol{I}_\text{mask}\!\in\!\mathbb{R}^{3\times h\times w}$, and a corresponding binary mask $\boldsymbol{M}\!\in\! \mathbb{R}^{1 \times h\times w}$ where $1$ represents the locations for invalid pixels and $0$ represents those for valid pixels. 
Considering that the global scene understanding usually involves multiple objects, in this work, we borrow a pretext multi-label classification model, denoted by $H$, to provide feature maps as the supervisions of learning the semantic priors. 
Unless otherwise specified, the pretext model is trained on the OpenImage dataset~\cite{kuznetsova2018open} with the asymmetric loss (ASL)~\cite{ben2020asymmetric}. Note that we do not fine-tune the model on any inpainting datasets used in this paper.
We hope the pretext model can provide more discriminative representations on specific objects, allowing the common knowledge to be transferred to the unsupervised image inpainting task.
To generate the supervisions of knowledge distillation, specifically, we up-sample the full image $\boldsymbol{I}_\text{full}\!\in\! \mathbb{R}^{3 \times h\times w}$ to obtain $\boldsymbol{I}_\text{full}^{\prime}\!\in\! \mathbb{R}^{3 \times 2h\times 2w}$, and then feed $\boldsymbol{I}_\text{full}^{\prime}$ to $H$ to obtain $L$ feature maps in total at multiple scales:
\begin{align}
    \big\{\boldsymbol{S}_\text{target}^{l}\big\}^{L}_{l=1} = H\big(\boldsymbol{I}_\text{full}^{\prime}\big), 
\end{align}
where $\boldsymbol{S}_\text{target}^{l}$ has a spatial size of $ \frac{h}{2^{l-1}}\times \frac{w}{2^{l-1}}$. %\revise{$L$ denotes the number of scales of the used semantic priors. In all experiments, we set $L=3$ and the spatial size in each scale is $256 \times 256$, $128 \times 128$, and $64 \times 64$, respectively.}
$L$ denotes the number of scales of the used semantic priors. In all experiments, we set $L=3$ and the spatial size in each scale is $256 \times 256$, $128 \times 128$, and $64 \times 64$, respectively.

At the beginning of the semantic prior learner, $L$ down-sampling layers are used, and a residual block%dense block~\cite{zhang2018residual} 
is applied after the last down-sampling layer to extract features from the masked image at different scales.
To align with the spatial resolution of $\boldsymbol{S}_\text{target}^{l}$, we also up-sample $\boldsymbol{I}_\text{mask}$ into $\boldsymbol{I}^{\prime}_\text{mask}\!\in\!\mathbb{R}^{3\times 2h\times 2w}$. The corresponding mask $\boldsymbol{M}$ is also up-sampled to $\boldsymbol{M}^\prime$ in the same manner, and concatenated with $\boldsymbol{I}^{\prime}_\text{mask}$ to form the input of the semantic prior learner:
\begin{align}
    \big\{\boldsymbol{S}^{l}_\text{enc}\big\}^{L}_{l=1} = \texttt{Enc}_\text{prior}\big(\boldsymbol{I}^{\prime}_\text{mask}, \  \boldsymbol{M}^{\prime}\big), 
\end{align}
where $\boldsymbol{S}^{l}_\text{enc}$ is the image representation learned from visible image contents that has the same spatial size as the semantic supervision $\boldsymbol{S}_\text{target}^{l}$. 
We here adopt the common practice about $\boldsymbol{M}$ in previous literature, keeping the region of $\boldsymbol{M}$ known during training and testing and focusing more on the recovery of missing content.
Next, we exploit multiple residual blocks and up-sampling layers to generate a pyramid of semantic priors with multi-scale skip connections:
\begin{equation}
\begin{split}
    \boldsymbol{S}^{L}_\text{prior} & = \texttt{Multi-ResBlock}\big(\boldsymbol{S}^{L}_\text{enc}\big),\\
    \boldsymbol{S}^{l}_\text{prior} & = \texttt{ResBlock}\big(U(\boldsymbol{S}^{l+1}_\text{prior}), \  \boldsymbol{S}^{l}_\text{enc}\big), \ l<L,
\end{split}
\end{equation}
where $U$ indicates the pixel-shuffle layer for up-sampling and $\{\boldsymbol{S}^{l}_\text{prior}\}_{l=1}^{L}$ indicates the \textit{prior pyramid}.
For each $\boldsymbol{S}^{l}_\text{prior}$, it has the spatial sizes of $\frac{h}{2^{l-1}}\times \frac{w}{2^{l-1}}$, and is followed by a distillation head $f^{l}$ of a $1\times1$ convolutional layer.
Finally, we apply a set of $\ell_1$ distillation losses to the deterministic semantic prior learner:
\begin{equation}
    % \revise{\mathcal{L}_{\text{prior}} = \sum_{l=1}^{L} \left\|\big(\boldsymbol{S}_\text{target}^{l}- f^{l}(\boldsymbol{S}^{l}_\text{prior})\big)\odot \big(\boldsymbol{1} + \alpha \boldsymbol{M}^{l}\big)\right\|_{1},}
    \mathcal{L}_{\text{prior}} = \sum_{l=1}^{L} \left\|\big(\boldsymbol{S}_\text{target}^{l}- f^{l}(\boldsymbol{S}^{l}_\text{prior})\big)\odot \big(\boldsymbol{1} + \alpha \boldsymbol{M}^{l}\big)\right\|_{1},
    \label{eq:distill}
\end{equation}
where $\odot$ denotes the Hadamard product, $\boldsymbol{M}^{l}$ is the resized mask with spatial size equal to $\boldsymbol{S}_\text{prior}^{l}$, and $\alpha$ is for the additional constraints on masked areas. 
The above objective functions are jointly used with final image reconstruction loss. In this way, the knowledge at different levels of the pretexts model can flow into the prior pyramid to best facilitate image inpainting.
Notably, the distillation heads $\{f^{l}\}_{l=1}^{L}$ and the pretext model $H$ are only used in the training phase and are all removed from SPN during testing.

\vspace{5pt}
\subsubsection{Learning stochastic prior pyramid}

\begin{figure}[t]
\begin{center}
\includegraphics[width=1.0\linewidth]{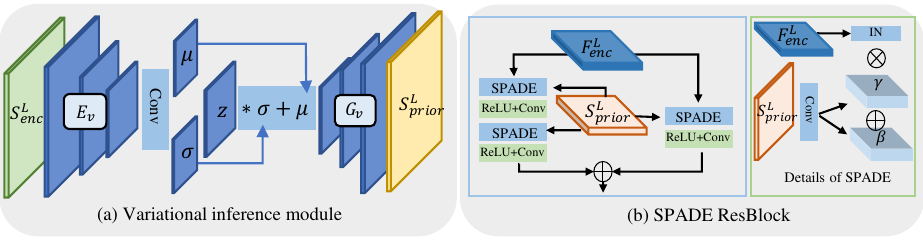}
\end{center}
\vspace{-1mm}
\caption{\textbf{Illustration of different sub-modules.} \textbf{(a) The variational inference module.} We use the re-parameterization where the latent variable $\boldsymbol{z}$ is sampled from a Gaussian distribution $\boldsymbol{z}\sim \mathcal N(\boldsymbol{0}, \boldsymbol{1})$. It is injected in the semantic prior learner, allowing diverse prior knowledge to be distilled from the pretext model. \textbf{(b) An example of the SPADE ResBlock}. It contains different numbers of SPADE modules~\protect\cite{park2019semantic} in separate branches to adaptively fuse the semantic priors $\boldsymbol{S}_\text{prior}^L$ and the image encoding features $\boldsymbol{F}_\text{enc}^L$. The two branches respond to different importance of integrating semantic priors in image restoration. 
}
% \vspace{-1mm}
\label{Variational_module}
\end{figure}

To resolve the contextual ambiguity, it is natural to consider various possibilities of the missing content in image inpainting.
Therefore, in addition to learning a prior pyramid, we further improve our previous work at IJCAI’21~\cite{DBLP:conf/ijcai/ZhangZTWCNWY21} by introducing a variational inference module in the semantic learner. 
This module is optional and handles the multi-modal distribution of possible missing content in the high-level feature space conditioned on the remaining part of the image.
By injecting this module in the semantic learner, we can easily extend SPN to probabilistic image inpainting, that is, generating diverse restoration results given an individual damaged image.
More specifically, we reformulate the semantic prior learner based on a Gaussian stochastic neural network. %~\cite{DBLP:conf/nips/SohnLY15}. 
In practice, as shown in \fig{Overview}(a), we simply embed a variational inference module between the high-level image representation $\boldsymbol{S}^{l}_\text{enc}$ and the corresponding learned priors $\boldsymbol{S}^{L}_\text{prior}$.
The details of this module are shown in \fig{Variational_module}(a).
It consists of an encoder $E_{v}$ that estimates the parameters of the posterior distribution of high-level semantics $q_{s}$ and a generator $G_{v}$ that produces diverse prior pyramids with the re-parameterization trick:
% \begin{equation}
% \begin{split}
%     \big[\boldsymbol{\mu}, \boldsymbol{\sigma}\big] &= E_{v}\big(\boldsymbol{S}^{L}_\text{enc}\big), \\
%     \revise{\boldsymbol{S}^{L}_\text{prior}} &\revise{= G_{v}\big(\boldsymbol{z} \odot \boldsymbol{\sigma} + \boldsymbol{\mu}\big), \  \boldsymbol{z}\sim \mathcal N\big(\boldsymbol{0}, \boldsymbol{1}\big).}
% \end{split}
% \end{equation}
\begin{equation}
\begin{split}
    \big[\boldsymbol{\mu}, \boldsymbol{\sigma}\big] = E_{v}\big(\boldsymbol{S}^{L}_\text{enc}\big);\  
    \boldsymbol{S}^{L}_\text{prior} = G_{v}\big(\boldsymbol{z} \odot \boldsymbol{\sigma} + \boldsymbol{\mu}\big), \  \boldsymbol{z}\sim \mathcal N\big(\boldsymbol{0}, \boldsymbol{1}\big).
\end{split}
\end{equation}

The multiple-level distillation losses are then used in the same way as the learning deterministic prior pyramid, which is shown in Eq. \eqref{eq:distill}.
We also optimize the KL divergence to constrain the distance between the learned posterior distribution and a prior distribution $p_{z}\sim \mathcal N(\boldsymbol{0}, \boldsymbol{1})$:
\begin{align}
    \mathcal{L}_{\text{KL}} = KL\big(q_{s} \ || \ p_{z}\big),
\end{align}
% \revise{where $q_{s}$ represents the learned posterior distribution.}
where $q_{s}$ represents the learned posterior distribution.

By optimizing the KL divergence within the semantic prior learner, our model is very different from existing approaches for stochastic image inpainting in the following perspective: \textit{It incorporates the distribution of potential semantics as a part of the learned priors, in the sense that different knowledge is distilled from the pretext model in a probabilistic way.}

% The KL divergence loss can also help the stabilization of the training process. Compared with other probabilistic inpainting methods which usually directly learn the mappings between pixels and noises, we model the distributions of the missing regions in a more semantic manner. By this means, the diversities of our method mainly come from the understanding of the global context information, which makes our model generate more pluralistic image structures.

\subsection{Fully Context-Aware Image Generator}
\label{context-aware_generator}

As shown in \fig{Overview}(b), we first exploit three convolutional layers to extract low-level visual representation from the masked image $\boldsymbol{I}_\text{mask}$ and mask $\boldsymbol{M}$: 
\begin{align}
    \boldsymbol{F}^{L}_\text{enc} = \texttt{Enc}_\text{img}\big(\boldsymbol{I}_\text{mask}, \boldsymbol{M}\big), 
\end{align}
where $\boldsymbol{F}^{L}_\text{enc}$ has the same spatial size of $\frac{h}{2^{L-1}}\times \frac{w}{2^{L-1}}$ as that of the learned semantic prior $\boldsymbol{S}^{L}_\text{prior}$.

% \begin{figure}[t]
% \begin{center}
% \includegraphics[width=0.8\linewidth]{pic/pic2_SPADE.pdf}
% \end{center}
% \vspace{-1mm}
% \caption{\textbf{An example of the SPADE ResBlock}. It contains different numbers of SPADE modules~\protect\cite{park2019semantic} in separate branches to adaptively fuse the semantic priors $\boldsymbol{S}_\text{prior}^L$ and the image encoding features $\boldsymbol{F}_\text{enc}^L$. The two branches respond to different importance of integrating semantic priors in image restoration. 
% }
% % \vspace{-1mm}
% \label{ResSPADE}
% \end{figure}

In the subsequent part of the image generator, we use the learned prior pyramid $\{\boldsymbol{S}^{l}_\text{prior}\}^{L}_{l=1}$ as semantic guidance to gradually refine the image encoding features $\boldsymbol{F}^{L}_\text{enc}$. 
However, we find it challenging to effectively combine these two branch of features into unified representations. 
First, since the prior pyramid is distilled from a pretext model that is learned in another domain, there may be significant domain conflicts between them. 
Second, the image encoding branch directly extracts low-level image features and focuses more on textures and local structure, while the prior pyramid branch focuses more on global scene understanding due to the effect of knowledge distillation from the multi-label classification model. 
As a result, we observe that fusing these two representation branch through concatenations will make them conflict, not only affecting local texture restoration but also increasing the difficulty of global context reasoning.

To adaptively incorporate the semantic priors into image encoding features and reduce the conflicts between these two branches, we exploit the \textit{spatially-adaptive normalization} module (SPADE)~\cite{park2019semantic} (see \fig{Variational_module}(b)) as the key component of the image generator. 
The SPADE module is originally proposed for image synthesis to maintain the spatial information of the semantic layout. 
Instead of using semantic maps or semantic layouts, in SPN, we take the learned prior pyramid along with the image encoding features as the inputs of SPADE, and perform non-parametric instance normalization on the image encoding features. 
Next, two sets of parameters $\boldsymbol{\gamma}$ and $\boldsymbol{\beta}$ are generated from the learned semantic priors to perform pixel-wise affine transformations on the image features. We here take the image feature $\boldsymbol{F}^{L}_\text{enc}$ as an example and have: 
% \begin{equation}
%     \begin{split}
%     \big[\boldsymbol{\gamma}, \  \boldsymbol{\beta}\big] &= \texttt{SPADE}\big(\boldsymbol{S}^{L}_\text{prior}\big), \\
%     \revise{\boldsymbol{F}^{\prime}} &\revise{= \boldsymbol{\gamma} \odot \texttt{IN}\big(\boldsymbol{F}^{L}_\text{enc}\big) + \boldsymbol{\beta},}
%     \end{split}
% \end{equation}
\begin{equation}
    % \begin{split}
    \big[\boldsymbol{\gamma}, \  \boldsymbol{\beta}\big] = \texttt{SPADE}\big(\boldsymbol{S}^{L}_\text{prior}\big); \ 
    \boldsymbol{F}^{\prime} = \boldsymbol{\gamma} \odot \texttt{IN}\big(\boldsymbol{F}^{L}_\text{enc}\big) + \boldsymbol{\beta},
    % \end{split}
\end{equation}
where $\boldsymbol{\gamma}, \boldsymbol{\beta}, \boldsymbol{F}^{L}_\text{enc}, \boldsymbol{F}^{\prime}\!\in\! \mathbb{R}^{c \times \frac{h}{2^{L-1}} \times \frac{w}{2^{L-1}}}$ and $c$ is the number of channels.
$\texttt{IN}(\cdot)$ indicates non-parametric instance normalization. 
In this way, the SPADE module provides adaptive refinements on the image encoding features guided by the prior pyramid, greatly facilitating context-aware image restoration.

Furthermore, we adopt the SPADE ResBlock from the work of Park \textit{et al.}~\cite{park2019semantic} to organize multiple SPADE modules in an imbalanced two-branch architecture, which uses different numbers of SPADE modules in different branches.
The basic idea is to \textit{make the model adaptively respond to the different importance of integrating semantic priors in the process of image generation.}
More specifically, we apply eight consecutive SPADE ResBlocks to the image features $\boldsymbol{F}^{L}_\text{enc}$ with the smallest spatial resolution to derive $\boldsymbol{F}^{L}_\text{dec}$, and one SPADE ResBlock to each level of image decoding features $\boldsymbol{F}^{l}_\text{dec}$ followed by an up-sampling layer ($U$). We obtain the final results $\widehat{\boldsymbol{I}}_\text{full}$ by 
\begin{equation}
    \begin{split}
    \boldsymbol{F}^{L}_\text{dec} &= \texttt{Multi-SPADEResBlock}\big(\boldsymbol{F}^{L}_\text{enc}, \  \boldsymbol{S}^{L}_\text{prior}\big),\\
    \boldsymbol{F}^{l}_\text{dec} &= \texttt{SPADEResBlock}\big(U(\boldsymbol{F}^{l+1}_\text{dec}), \  \boldsymbol{S}^{l}_\text{prior}\big), \ l<L,\\
    \widehat{\boldsymbol{I}}_\text{full} &= \texttt{Conv}\big(\boldsymbol{F}^{l}_\text{dec}\big), \ l=1.
    \end{split}
\label{eq:ResSPADE}
\end{equation}

Different from our previous work~\cite{DBLP:conf/ijcai/ZhangZTWCNWY21}, we make the image generator \textit{fully context-aware} by leveraging the SPADE ResBlocks to adaptively and progressively incorporate the learned feature pyramid of multi-scale priors. The new architecture of the image generator eases the joint training of global contextual reasoning and local texture completion.

\subsection{Final Objective Functions}
\label{loss_function}
We exploit multi-scale semantic supervisions from the pretext model and jointly optimize the prior learner and the fully context-aware image generator.
SPN can alternatively handle deterministic or probabilistic image inpainting by using or not using the variational inference module in the prior learner.
In this part, we introduce the overall loss functions under each setup separately.

\vspace{5pt}
\subsubsection{Losses for deterministic inpainting}

We use the reconstruction and adversarial losses to train SPN for deterministic inpainting. The reconstruction loss is in an $\ell_1$ form and constrains the pixel-level distance between the ground truth images and the inpainting results, with more attention on the missing content:
\begin{align}
    \mathcal{L}_{\text{img}} = \left\|\big(\boldsymbol{I}_\text{full} - \widehat{\boldsymbol{I}}_\text{full}\big) \odot \big(\boldsymbol{1} + \delta \boldsymbol{M}\big)\right\|_{1}.
\end{align}

The adversarial loss, as adopted in existing approaches~\cite{yu2020region,nazeri2019edgeconnect}, is used to encourage realistic inpainting results:
\begin{equation}
    \begin{split}
     \mathcal{L}_{\text{adv}} &= -\mathbb{E}_{\widehat{I}}  \Big[\log\big(1-D(\boldsymbol{\widehat{I}}_\text{full})\big)\Big],\\
    \mathcal{L}^{D}_{\text{adv}} &= \mathbb{E}_{I} \Big[\log D\big(\boldsymbol{I}_\text{full}\big)\Big] +
    \mathbb{E}_{\boldsymbol{\widehat{I}}} \Big[\log\big(1-D(\widehat{\boldsymbol{I}}_\text{full})\big)\Big],
    \end{split}
\end{equation}
where $D$ represents the discriminator and $\mathcal{L}^{D}_{\text{adv}}$ encourages the discriminator to distinguish real and generated images.
By combining $\mathcal{L}_{\text{adv}}$ with the aforementioned loss terms, the full objective function can be written as
\begin{align}
    \mathcal{L}_\text{DET} = \mathcal{L}_{\text{prior}} +  \lambda_{1}\mathcal{L}_{\text{img}} + \lambda_{2}\mathcal{L}_{\text{adv}},
    \label{eq:deter_loss}
\end{align}
where the form of $\mathcal{L}_{\text{prior}}$ is shown in Eq. \eqref{eq:distill}. %\revise{The grid search range for $\lambda_{1}$ is $[0.1, 1, 10]$ and the range for $\lambda_{2}$ is same as $\lambda_{1}$.} We finally set $\lambda_{1}=10$, $\lambda_{2} = 1$, and $\delta = 4$. Similarly, we set $\alpha = 3$ in Eq. \eqref{eq:distill}.
The grid search range for $\lambda_{1}$ is $[0.1, 1, 10]$ and the range for $\lambda_{2}$ is same as $\lambda_{1}$. We finally set $\lambda_{1}=10$, $\lambda_{2} = 1$, and $\delta = 4$. Similarly, we set $\alpha = 3$ in Eq. \eqref{eq:distill}.

\vspace{5pt}
\subsubsection{Losses for probabilistic inpainting}
% Compared with the deterministic inpainting setup, the probabilistic image inpainting focuses on generating diverse realistic results that are semantically consistent with the remaining image content. 
% %
% Therefore, 
Except for the loss funtions in Eq. \eqref{eq:deter_loss}, we further use the feature matching loss, the perceptual loss and the perceptual diversity loss~\cite{DBLP:conf/cvpr/LiuWHSH021} to train the entire model for probabilistic inpainting.
%~\cite{DBLP:conf/cvpr/Wang0ZTKC18} ~\cite{DBLP:conf/eccv/JohnsonAF16}

The feature matching loss %can stabilize the training process of conditional generative models. It 
measures the distance between feature maps extracted from $N$ layers of the discriminator given real and generated images. We use  $\phi^{i}$ to denote the mapping function from the input image to the feature map at the $i$-th layer in the discriminator. 
For the perceptual loss, we use a fixed VGG19 model pre-trained on ImageNet to extract multi-scale visual representations and reduce their distance in the semantic space. % by minimizing the perceptual loss. ~\cite{DBLP:journals/corr/SimonyanZ14a}
We use $\varphi^{i}$ to denote the feature map from the $i$-th layer in the VGG19 model and $K$ is the total number of layers we use.
Combined with the feature matching loss that has similar forms, the two loss terms can be summarized as
\begin{equation}
    \begin{split}
     \mathcal{L}_{\text{feature}} = \ & \frac{1}{N}\sum_{i=1}^{N}\left\|\phi^{i}\big(\boldsymbol{I}_\text{full}\big) - \phi^{i}\big(\widehat{\boldsymbol{I}}_\text{full}\big)\right\|_{1} + \\ 
     &\frac{1}{K}\sum_{i=1}^{K}\left\|\varphi^{i}\big(\boldsymbol{I}_\text{full}\big) - \varphi^{i}\big(\widehat{\boldsymbol{I}}_\text{full}\big)\right\|_{1}.
    \end{split}
    \label{eq:match}
\end{equation}

As for the perceptual diversity loss, it encourages SPN to produce semantically diverse results in the missing areas. It is also conducted on the feature maps from the pre-trained VGG19 model:
\begin{align}
    \mathcal{L}_{\text{diverse}} = \frac{1}{K}\sum_{i=1}^{K}\frac{1}{\big\|\varphi^{i}(\widehat{\boldsymbol{I}}_{1})\cdot \boldsymbol{M}_{i} - \varphi^{i}(\widehat{\boldsymbol{I}}_{2})\cdot \boldsymbol{M}_{i}\big\|_{1} + \epsilon},
    \label{eq:diverse}
\end{align}
where $\widehat{\boldsymbol{I}}_{1}$ and $\widehat{\boldsymbol{I}}_{2}$ denote restored images generated from the same damaged image but with different noises, $\boldsymbol{M}_{i}$ is the resized mask which has the same spatial size as $\varphi^{i}$, and $\epsilon$ is the perturbation term to avoid the outlier. The full objective function for probabilistic image inpainting can be written as
\begin{align}
    \mathcal{L}_\text{PROB} = \mathcal{L}_\text{DET} + \lambda_{3}\mathcal{L}_\text{feature} + \lambda_{4}\mathcal{L}_\text{diverse} + \lambda_{5}\mathcal{L}_\text{KL},
\end{align}
The grid search range for both $\lambda_{3}$ and $\lambda_{4}$ is $[0.1, 1, 10]$ and the range for $\lambda_{5}$ is $[0.01, 0.05, 0.1]$. Finally, we have $\lambda_{3}=10$, $\lambda_{4}=1$, and $\lambda_{5}=0.05$. We refer to SPADE~\cite{park2019semantic} to set other parameters including $K=5$, $N=3$, and $\epsilon=1e^{-5}$.
% \revise{The grid search range for both $\lambda_{3}$ and $\lambda_{4}$ is $[0.1, 1, 10]$ and the range for $\lambda_{5}$ is $[0.01, 0.05, 0.1]$}. Finally, we have $\lambda_{3}=10$, $\lambda_{4}=1$, and $\lambda_{5}=0.05$. \revise{We refer to SPADE~\cite{park2019semantic} to set other parameters including $K=5$, $N=3$, and $\epsilon=1e^{-5}$.}%The number of selected layers ($N, K$) in Eq. \eqref{eq:match}-\eqref{eq:diverse} are the same as those in the work of SPADE~\cite{park2019semantic}.
%%%%%%%%%%%%%%%%%%%%%%%%%%%%%%%%%%%%%%%%%%%%%%%%%%%%%%%%%%%%%%%%%%%%
% experiments
%%%%%%%%%%%%%%%%%%%%%%%%%%%%%%%%%%%%%%%%%%%%%%%%%%%%%%%%%%%%%%%%%%%%

\section{Experiments}
\label{experiments}

\subsection{Experimental Setups} 

\subsubsection{Datasets}

We validate the effectiveness of SPN on four datasets in total. 
Under the deterministic image inpainting setup, we evaluate SPN on three datasets. 
For the probabilistic inpainting setup, we use the CelebA-HQ~\cite{DBLP:conf/iclr/KarrasALL18} and Paris StreetView dataset~\cite{doersch2015makes}.%, where diverse results can be generated given each individual damaged input image and the random latent variables drawn from Gaussian distribution.
Furthermore, we use an external dataset to provide irregular masks.
In both training and testing phases, for CelebA, we follow the common practice~\cite{yu2020region, li2020recurrent} to perform center crop on images and then resize them to $256 \times 256$. For other datasets, raw images are resized to $256 \times 256$ directly.
\begin{itemize}%[leftmargin=*]
    \vspace{-5pt}
    \item \textbf{Places2}~\cite{zhou2017places}. It is one of the most challenging datasets containing over $1.8$ million images from $365$ different scenes. We use the standard training set with $1.8$ million images for training and perform evaluations on the first $12{,}000$ images in the validation set.
    \vspace{-5pt}
    \item \textbf{CelebA}~\cite{liu2015deep}. It contains about $200{,}000$ diverse celebrity facial images. we use the original training set that includes over $162{,}000$ images for training and use the first $12{,}000$ images in the test set for testing.
    \vspace{-5pt}
    \item 
    \textbf{Paris StreetView}~\cite{doersch2015makes}. It collects images from street views of Paris and mainly contains different highly structured facades. We use $14{,}900$ images for training and $100$ images for testing. We use this dataset under both deterministic and probabilistic setups.
    \vspace{-5pt}
    \item \textbf{CelebA-HQ}~\cite{DBLP:conf/iclr/KarrasALL18}. It is a human facial dataset used by previous probabilistic inpainting methods~\cite{zheng2019pluralistic,DBLP:conf/cvpr/LiuWHSH021}. It contains a higher quality of texture detail than CelebA. Following the existing work~\cite{DBLP:conf/cvpr/LiuWHSH021}, we use about $27{,}000$ images for training and $2{,}800$ images for testing.
    \vspace{-5pt} 
    \item 
    \textbf{Irregular masks}~\cite{liu2018image}. This dataset is used to provide irregular masks in our experiments. As a common practice~\cite{yu2020region}, we perform random flipping at training time. %At test time, the masks are randomly sampled from the test set. %, so that they are different from training. 
    A total number of $12{,}000$ irregular masks are grouped by six intervals of the mask ratio: $0\%$-$10\%$, $10\%$-$20\%$, $\ldots$, $50\%$-$60\%$.
\end{itemize}

% \vspace{5pt}
\subsubsection{Compared methods}
We extend our previous work by providing stronger baseline models.
For the deterministic setup, we compare SPN with RFR~\cite{li2020recurrent}, RN~\cite{yu2020region}, MFE~\cite{DBLP:conf/eccv/LiuJSHY20}, EC~\cite{nazeri2019edgeconnect}, WF~\cite{yu2021wavefill}, and CTSDG~\cite{guo2021image}. 
Particularly, we denote our previous work as SPL~\cite{DBLP:conf/ijcai/ZhangZTWCNWY21}, short for the Semantic Prior Learner.
%
% Among these methods, RN introduces the region normalization module to separate the normalization processes between the valid and invalid regions. RFR introduces a recurrent reasoning framework to perform confidence guided recurrent restoration in missing areas. \revise{WF uses wavelet transformation to explicitly restore image contents in different frequency bands.} These approaches mainly focus on local texture consistency. 
Among these methods, RN, RFR, and WF introduce different network architectures to improve both encoding and decoding stages.
%
% On the other hand, EC adopts a two-stage generation framework to first generate edge maps as the structure guidance. \revise{CTSDG follows this line and proposes a dual generation pipeline.} MFE directly exploits smooth images as additional structural supervisions to help image completion. These two methods use external guidance about the scene, but still suffer from distorted structures due to the lack of global semantic understanding. 
On the other hand, EC, CTSDG, and MFE try to use edge maps or smooth images as additional structural supervisions to help image completion. However, they still suffer from distorted structures due to the lack of global semantic understanding.
For probabilistic inpainting, we compare SPN with PIC~\cite{zheng2019pluralistic} and DSI~\cite{DBLP:conf/cvpr/Peng0XL21}.
%
%PIC introduces a two-branches VAE framework for diverse image completion.
%
%DSI uses a multi-stages auto-regressive framework. %based on VQ-VAE~\cite{DBLP:conf/nips/OordVK17}. 
%For fair comparisons, we directly evaluate these methods with the available pre-trained models. Otherwise, we re-train their models with the provided code.

% \vspace{5pt}
\subsubsection{Implementation details}
In our experiments, we extract feature maps at three scales from the multi-label classification model to form the distillation targets. The classification model was pre-trained on the OpenImage dataset~\cite{kuznetsova2018open} with the asymmetric loss (ASL)~\cite{ben2020asymmetric}, and is not fine-tuned on any datasets used in this work. Besides, we also use the patch-based discriminator for adversarial training as in the previous work~\cite{yu2020region,DBLP:conf/eccv/LiuJSHY20}. 
Our model is trained by Adam solver with $\beta_1=0.0$ and $\beta_2=0.9$. The initial learning rate is set to $1e^{-4}$ for all experiments, and we decay the learning rate to $1e^{-5}$ in the last quarter of the training process. %Take the Paris StreetView dataset as an example, we train SPN for $150K$ iterations and decay the learning rate at about $127K$ iterations.
%
%We increase the total training iterations according to the dataset volume. The batch size for all experiments is $8$. 
%
SPN is trained on two V100 GPUs, and it takes about one day to train our model on Paris StreetView dataset. For more details, please refer to our code and pre-trained models at \url{https://github.com/WendongZh/SPN}.

\begin{table}[t!]
% \small WF~\cite{yu2021wavefill}, and CTSDG~\cite{guo2021image}
\footnotesize
\renewcommand{\tabcolsep}{2.0pt}
% \footnotesize
    \caption{Quantitative results on the three datasets under the deterministic image inpainting setup. Note that SPL \cite{DBLP:conf/ijcai/ZhangZTWCNWY21} is our previous work at IJCAI 2021.
	}
	%\vspace{-3mm}
	\label{tab:Quantitative_comparison in datases}
	\centering
	\begin{tabular}{l|l|ccc|ccc|ccc}
		\hline
		\multicolumn{2}{c|}{Dataset} & \multicolumn{3}{c|}{Places2} & \multicolumn{3}{c|}{CelebA} & \multicolumn{3}{c}{Paris StreetView}\\
		\hline
		\multicolumn{2}{c|}{Mask Ratio} & 0.2-0.4 & 0.4-0.6 & All & 0.2-0.4 & 0.4-0.6 & All & 0.2-0.4 & 0.4-0.6 & All  \\
		\hline
		\hline
		\multirow{8}{*}{SSIM$^{\uparrow}$} 
	    &EC~\cite{nazeri2019edgeconnect}      & 0.847 & 0.695 & 0.832 & 0.922 & 0.815 & 0.905 & 0.880 & 0.740 & 0.849\\
		&MFE~\cite{DBLP:conf/eccv/LiuJSHY20}     & 0.816 & 0.652 & 0.804 & 0.916 & 0.807 & 0.900 & 0.872 & 0.707 & 0.834\\
		&RFR~\cite{li2020recurrent}     & 0.856 & 0.704 & 0.839 & 0.931 & 0.838 & 0.917 &  0.893 & 0.763 & 0.863\\
		&RN~\cite{yu2020region}      & 0.875 & 0.726 & 0.855 & 0.941 & 0.852 & 0.925 & 0.891 & 0.756 & 0.861\\
		&\revise{WF}~\cite{yu2021wavefill}      & \revise{0.855} & \revise{0.647} & \revise{0.820} & - & - & - & \revise{0.895} & \revise{0.765} & \revise{0.864}\\
		&\revise{CTSDG}~\cite{guo2021image}      & \revise{0.848} & \revise{0.696} & \revise{0.837} & \revise{0.923} & \revise{0.819} & \revise{0.907} & \revise{0.888} & \revise{0.746} & \revise{0.855}\\
		&SPL~\cite{DBLP:conf/ijcai/ZhangZTWCNWY21}     & 0.894 & 0.759 & 0.875 & 0.950 & 0.869 & 0.935 & 0.911 & 0.790 & 0.882\\
		&Ours      & \textbf{0.897} & \textbf{0.763} & \textbf{0.877} & \textbf{0.952} & \textbf{0.871} & \textbf{0.937} & \textbf{0.916} & \textbf{0.795} & \textbf{0.886}\\
		\hline
		\multirow{8}{*}{PSNR$^{\uparrow}$} 
		&EC~\cite{nazeri2019edgeconnect}      & 24.17 & 20.32 & 25.07 & 29.47 & 24.08 & 30.34 & 27.65 & 22.81 & 27.50\\
		&MFE~\cite{DBLP:conf/eccv/LiuJSHY20}     & 22.89 & 19.13 & 23.94 & 29.03 & 23.85 & 30.13 & 27.22 & 22.07 & 27.07\\
		&RFR~\cite{li2020recurrent}     & 24.38 & 20.42 & 25.48 & 30.33 & 25.09 & 31.38 & 28.32 & 23.71 & 28.36\\
		&RN~\cite{yu2020region}      & 25.44 & 21.08 & 26.37 & 31.17 & 25.51 & 32.17 & 28.44 & 23.53 & 28.37\\
		&\revise{WF}~\cite{yu2021wavefill}      & \revise{24.57} & \revise{19.31} & \revise{24.77} & - & - & - & \revise{28.66} & \revise{23.88} & \revise{28.21}\\
		&\revise{CTSDG}~\cite{guo2021image}      & \revise{24.16} & \revise{20.36} & \revise{25.54} & \revise{29.48} & \revise{24.21} & \revise{30.62} & \revise{28.08} & \revise{23.39} & \revise{28.11}\\
		&SPL~\cite{DBLP:conf/ijcai/ZhangZTWCNWY21}     & 26.47 & 22.05 & 27.55 & 32.17 & 26.43 & 33.23 & 29.34 & 24.47 & 29.38\\
		&Ours    & \textbf{26.65} & \textbf{22.18} & \textbf{27.72} & \textbf{32.39} & \textbf{26.54} & \textbf{33.52} & \textbf{29.76} & \textbf{24.67} & \textbf{29.69}\\
		%&Mean $\mathit{l}_1^{\dag}$ & 0.1-0.2 & 0.3-0.4 & 0.5-0.6 & 0.1-0.2 & 0.3-0.4 & 0.5-0.6 & 0.1-0.2 & 0.3-0.4 & 0.5-0.6\\
		\hline
		\multirow{8}{*}{\shortstack{MAE$^{\downarrow}$ \\ ($\times 10^{-1}$)}} 
		&EC~\cite{nazeri2019edgeconnect}       & 0.221 & 0.456 & 0.248 & 0.111 & 0.274 & 0.140 & 0.148 & 0.364 & 0.204\\
		&MFE~\cite{DBLP:conf/eccv/LiuJSHY20}       & 0.266 & 0.542 & 0.296 & 0.119 & 0.284 & 0.145 & 0.159 & 0.412 & 0.226\\
		&RFR~\cite{li2020recurrent}       & 0.212 & 0.445 & 0.240 & 0.099 & 0.237 & 0.121 & 0.134 & 0.324 & 0.182\\
		&RN~\cite{yu2020region}        & 0.187 & 0.412 & 0.218 & 0.091 & 0.229 & 0.115 & 0.136 & 0.337 & 0.187\\
		&\revise{WF}~\cite{yu2021wavefill}      & \revise{0.216} & \revise{0.583} & \revise{0.285} & - & - & - & \revise{0.132} & \revise{0.330} & \revise{0.185}\\
		&\revise{CTSDG}~\cite{guo2021image}      & \revise{0.220} & \revise{0.452} & \revise{0.237} & \revise{0.109} & \revise{0.267} & \revise{0.135} & \revise{0.138} & \revise{0.338} & \revise{0.189}\\
		&SPL~\cite{DBLP:conf/ijcai/ZhangZTWCNWY21}       & 0.164 & 0.363 & 0.191 & 0.080 & 0.203 & 0.102 & 0.115 & 0.292 & 0.161\\
		&Ours      & \textbf{0.158} & \textbf{0.351} & \textbf{0.185} & \textbf{0.077} & \textbf{0.200}& \textbf{0.099} & \textbf{0.109} & \textbf{0.282} & \textbf{0.154}\\
		\hline
        \multirow{8}{*}{FID$^{\downarrow}$} 
		&EC~\cite{nazeri2019edgeconnect}       & 9.63 & 23.04 & 6.08 & 10.41 & 14.40 & 7.24 & 43.45 & 87.70 & 44.95\\
		&MFE~\cite{DBLP:conf/eccv/LiuJSHY20}      & 21.07 & 48.74 & 15.38 & 11.12 & 15.42 & 7.95 & 47.22 & 100.85 & 50.33\\
		&RFR~\cite{li2020recurrent}      & 8.43 & 21.82 & 5.27 & 10.22 & 14.57 & 7.15 & 37.51 & 77.15 & 39.44\\
		&RN~\cite{yu2020region}       & 9.27 & 25.00 & 6.10 & 11.39 & 18.00 & 8.16 & 64.78 & 134.00 & 66.54\\
		&\revise{WF}~\cite{yu2021wavefill}      & \revise{7.75} & \revise{42.46} & \revise{9.37} & - & - & - & \textbf{\revise{33.20}} & \revise{73.31} & \revise{38.36}\\
		&\revise{CTSDG}~\cite{guo2021image}      & \revise{15.88} & \revise{42.19} & \revise{12.36} & \revise{11.32} & \revise{17.57} & \revise{8.14} & \revise{46.52} & \revise{98.76} & \revise{49.19}\\
		&SPL~\cite{DBLP:conf/ijcai/ZhangZTWCNWY21}      & 8.30 & 23.98 & 5.56 & 10.27 & 12.94 & 6.88 & 41.41 & 93.41 & 46.13\\
		&Ours     & \textbf{7.68} & \textbf{20.73} & \textbf{4.73} & \textbf{10.16} & \textbf{12.38} & \textbf{6.77} & 34.00 & \textbf{70.76} & \textbf{36.82}\\
		\hline
	\end{tabular}
	%\vspace{-2mm}
\end{table}

\subsection{Deterministic Image Inpainting}
\begin{figure*}[t!]
\centering
\includegraphics[width=1.0\linewidth]{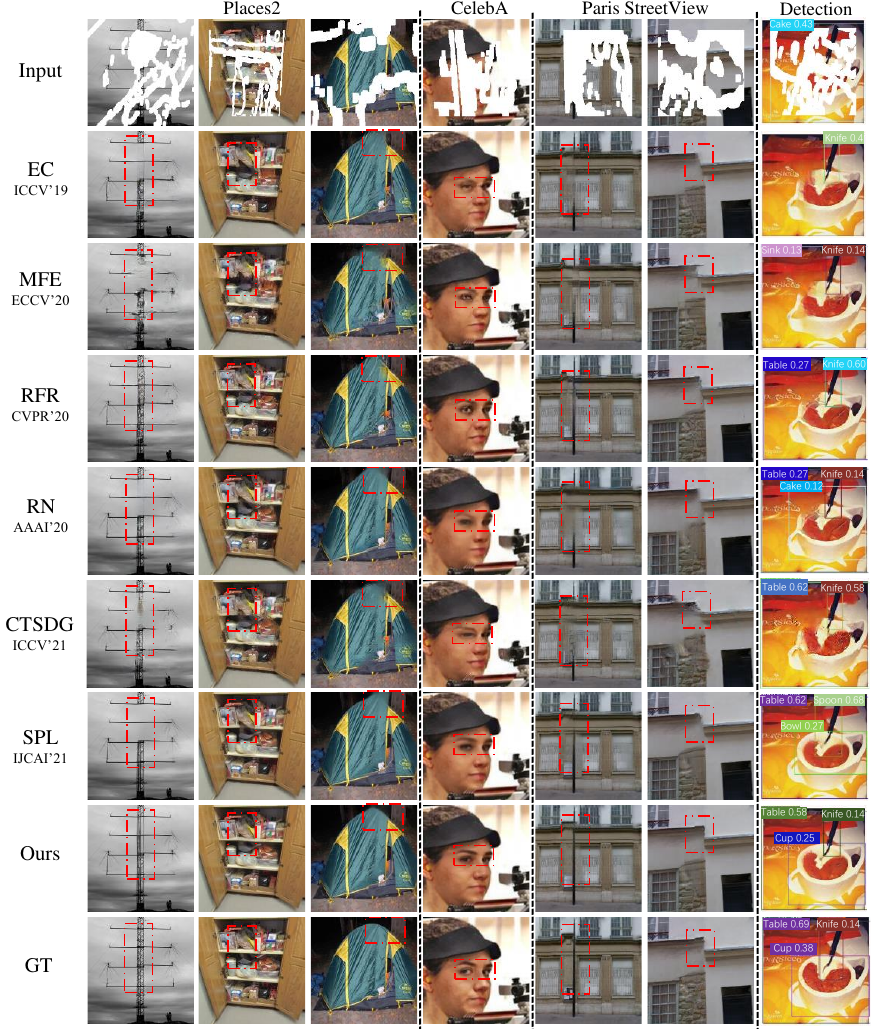}
% \includegraphics[width=1.0\linewidth]{pic/result1_v3_v2_hl.pdf}
%\vspace{-5mm}
\caption{Qualitative results of deterministic image inpainting. In the ``\textit{Detection}'' column, we draw detection results on the generated images in Place2 to show that SPN provides a better understanding of object-level image content and scene structure. SPL \cite{DBLP:conf/ijcai/ZhangZTWCNWY21} is our previous work at IJCAI 2021.}
%\vspace{-3mm}
\label{Qualititative}
\end{figure*}

We first provide both quantitative and qualitative comparison results under the deterministic inpainting setup in Table~\ref{tab:Quantitative_comparison in datases} and Figure~\ref{Qualititative}, respectively. 
For quantitative comparisons, we use the structural similarity (SSIM), peak signal-to-noise ratio (PSNR), and mean $\ell_{1}$ error (MAE) as evaluation metrics, which are commonly used in previous works~\cite{yu2020region,li2020recurrent}. 
Besides, we also use Fréchet inception distance (FID)~\cite{DBLP:conf/nips/HeuselRUNH17} as the human perception-level metric. %, which exploits deep visual representations extracted from a pre-trained neural network and measures the distance between the real and generated images in a perceptual view. 
For each dataset, we first randomly select mask images from the entire irregular mask dataset to obtain the mask set for evaluation. The mask selections are performed independently between different image datasets, and the selected masks are randomly assigned to test set images to form the mask-image pairs. For the same image dataset, the mask-image pairs are held for different methods to obtain fair comparison results. 
Since the mask dataset contains mask images with different mask ratios, we also show detailed quantitative results within different mask intervals.
Specifically, as shown in Table~\ref{tab:Quantitative_comparison in datases}, the results in columns denoted by ``ALL'' are obtained on the entire testing set; In other columns, the results are obtained throughout mask-images pairs within corresponding mask intervals.

% \vspace{5pt}
\subsubsection{Quantitative comparisons}
We have the following observations from Table~\ref{tab:Quantitative_comparison in datases}. \textbf{First}, the proposed SPN achieves the best results in most evaluation metrics on all three datasets. %with different mask intervals. 
It not only outperforms its previous version SPL, but also obtains significant improvements (up to $1$dB for PSNR) compared with other methods. \textbf{Second}, the improvements from our model are consistent across the three datasets. %Note that the difficulty of deterministic image inpainting is closely related to the complexity of image datasets and mask ratios. For example, Places2 is the most challenging dataset since it contains diverse scenes, which makes all compared models achieve the lowest performance. However, 
Particularly, SPN obtains the most significant improvements over other competitors such as EC and MFE on the Place2 dataset, which demonstrates that SPN can handle more difficult image scenarios. \textbf{Third}, for FID, our model greatly improves the results of its previous work SPL, achieving the best performance among all compared models. This improvement indicates the effectiveness of the newly proposed multi-scale prior learner and fully context-aware image generator in better capturing global context semantics.

\begin{figure*}[t!]
\centering
\includegraphics[width=1.0\linewidth]{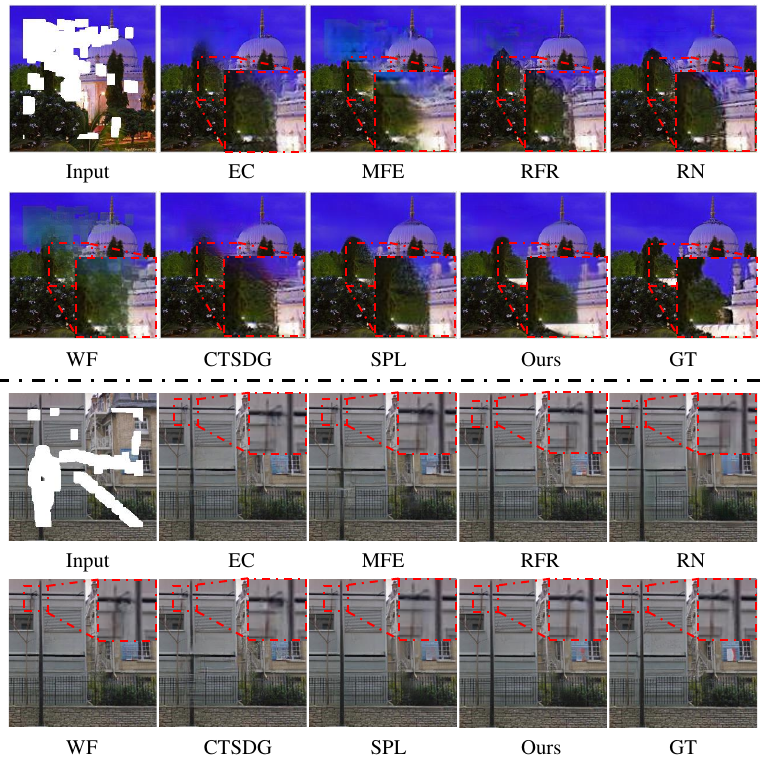}
%\vspace{-5mm}
\caption{\revise{Showcases with highlighted details. \textbf{Upper:} samples from the Place2 dataset. \textbf{Lower:} samples from the Paris StreetView dataset.}}
%\vspace{-3mm}
\label{fig:details}
\end{figure*}

% \vspace{5pt}
\subsubsection{Qualitative comparisons}
\fig{Qualititative} gives the showcases of the restored images on three datasets. 
For the Places2 dataset which contains complex scenes and various objects, our model successfully generates more complete global and local structures for missing regions. %Take the first column as an example, SPN restores clear global structures for the masked antenna. In contrast, other methods except for RFR all fail in this case, including our previous work SPL. 
In other columns, we can see that the results of RFR still suffer from distorted structures, while SPN can produce more complete structures as well as clearer details. %We have similar observations on the Paris StreetView dataset. 
Particularly, for the last column, we perform object detection on the restored images to show whether the compared models generate images by effectively understanding the semantics of the scene.
We can see that, compared with other methods, the generated images from SPN lead to correct object detection results with higher confidence scores. 

\revise{We also provide more samples with highlighted details in Fig~\ref{fig:details}. We can observe that results from SPN contain more clear and sharp local details, which further demonstrate the effectiveness of the proposed semantic pyramid.}

\subsection{Probabilistic Image Inpainting}

%In this part, we train SPN on the CelebA-HQ dataset under the probabilistic inpainting setup, and compare it with existing probabilistic inpainting methods in Table~\ref{tab:Quantitative_comparison on CelebA-HQ} and \fig{Qualitative_celebahq}.

In this part, we train SPN for probabilistic image inpainting and compare with existing probabilistic approaches on CelebA-HQ and Paris StreetView. 

\subsubsection{Quantitative comparisons}
The evaluation process in probabilistic inpainting is similar to that in deterministic inpainting. Since probabilistic models can generate diverse plausible results given a single input image, we select images with the best PSNR performance from $5$ different samples to calculate the quantitative results.
%
%Specifically, for each input image, we draw $5$ latent variables randomly and use them to generate corresponding image inpainting results. 
%
%\textcolor{blue}{We also report the performance of our previous deterministic work SPL~\cite{DBLP:conf/ijcai/ZhangZTWCNWY21} for reference.}
%
From Table~\ref{tab:Quantitative_comparison on CelebA-HQ}, % and Table~\ref{tab:Quantitative_comparison on Paris probabilistic}, 
we can observe that SPN significantly outperforms other probabilistic inpainting approaches in all evaluation metrics on different datasets.
Although SPN underperforms our previous deterministic approach SPL~\cite{DBLP:conf/ijcai/ZhangZTWCNWY21} on most metrics, it still achieves better results on FID with mask ratio $40\%-60\%$. This result can be explained from two aspects. First, SPN pays extra capacities to learn the multi-model distribution of the generated image, which results in the performance degeneration especially in pixel-level metrics. Second, the learned prior pyramid provides consistent multi-scale semantic guidance, which contributes to the restoration of large missing areas.

% \vspace{5pt}
\subsubsection{Qualitative comparisons}

\begin{table}[t!]
\caption{Quantitative results of probabilistic image inpainting on the CelebA-HQ dataset, which is one of the most commonly used benchmarks for evaluating probabilistic inpainting models. We also provide results on the Paris StreetView dataset for further comparison.
%\textcolor{red}{Delete results from SPL?} %Results from SPL~\cite{DBLP:conf/ijcai/ZhangZTWCNWY21} are calculated under the deterministic setup. 
}
%\vspace{-3mm}
\label{tab:Quantitative_comparison on CelebA-HQ}
% \small
\footnotesize
\renewcommand{\tabcolsep}{2.0pt}
\centering
% \resizebox{\columnwidth}{!}{
\begin{tabular}{l|l|cccc|cccc}
	\hline
	\multicolumn{2}{c|}{Dataset} &\multicolumn{4}{c|}{CelebA-HQ} &\multicolumn{4}{c}{Paris StreetView}\\
	\hline
	\multicolumn{2}{c|}{Mask Ratio} & 0.0-0.2 & 0.2-0.4 & 0.4-0.6 &All & 0.0-0.2 & 0.2-0.4 & 0.4-0.6 &All \\
	\hline
	\hline
	\multirow{3}{*}{SSIM$^{\uparrow}$} 
	&PIC~\cite{zheng2019pluralistic}     & 0.962  & 0.875 & 0.742 & 0.860 & 0.944  & 0.844 & 0.660 & 0.802\\
	&DSI~\cite{DBLP:conf/cvpr/Peng0XL21}      & 0.964  & 0.880 & 0.758 & 0.868 & -  & - & - & -\\
	%&SPL~\cite{DBLP:conf/ijcai/ZhangZTWCNWY21}      & 0.973  & 0.906 & 0.802 & 0.894\\
	%&Ours      & \textbf{0.973} & \textbf{0.901} & \textbf{0.788} & \textbf{0.888}\\
	&Ours      & \textbf{0.971} & \textbf{0.894} & \textbf{0.777} & \textbf{0.881} & \textbf{0.966}  & \textbf{0.896} & \textbf{0.763} & \textbf{0.865}\\
% 	\hline
    %\cline{2-6}
	%&SPL~\cite{DBLP:conf/ijcai/ZhangZTWCNWY21}      & 0.973  & 0.906 & 0.802 & 0.894\\
	\hline
	\multirow{3}{*}{PSNR$^{\uparrow}$} 
	&PIC~\cite{zheng2019pluralistic}      & 33.80  & 26.56 & 22.13 & 27.49 & 31.96  & 26.16 & 21.28 & 25.93\\
	&DSI~\cite{DBLP:conf/cvpr/Peng0XL21}      & 34.42 & 26.92 & 22.75 & 28.03 & -  & - & - & -\\
	%&SPL~\cite{DBLP:conf/ijcai/ZhangZTWCNWY21}      & 36.05  & 28.38 & 24.18 & 29.53\\
	%&Ours       & \textbf{36.11}  & \textbf{28.15} & \textbf{23.70} & \textbf{29.32}\\
	&Ours       & \textbf{35.72}  & \textbf{27.68} & \textbf{23.32} & \textbf{28.90} & \textbf{35.03}  & \textbf{28.71} & \textbf{23.66} & \textbf{28.56}\\
	%&Mean $\mathit{l}_1^{\dag}$ & 0.1-0.2 & 0.3-0.4 & 0.5-0.6 & 0.1-0.2 & 0.3-0.4 & 0.5-0.6 & 0.1-0.2 & 0.3-0.4 & 0.5-0.6\\
% 	\hline
    %\cline{2-6}
	%&SPL~\cite{DBLP:conf/ijcai/ZhangZTWCNWY21}      & 36.05  & 28.38 & 24.18 & 29.53\\
	\hline
	\multirow{3}{*}{\shortstack{MAE$^{\downarrow}$ \\ ($\times 10^{-1}$)}} 
	&PIC~\cite{zheng2019pluralistic}       & 0.046  & 0.161 & 0.359 & 0.187 & 0.067  & 0.185 & 0.473 & 0.261\\
	&DSI~\cite{DBLP:conf/cvpr/Peng0XL21}       & 0.044 & 0.155 & 0.329 & 0.175 & -  & - & - & -\\
	%&SPL~\cite{DBLP:conf/ijcai/ZhangZTWCNWY21}      & 0.0036  & 0.0126 & 0.0269 & 0.0143\\
	%&Ours         & \textbf{0.0036}  & \textbf{0.0133} & \textbf{0.0292} & \textbf{0.0153}\\
	&Ours         & \textbf{0.038}  & \textbf{0.140} & \textbf{0.305} & \textbf{0.160} & \textbf{0.045}  & \textbf{0.127} & \textbf{0.323} & \textbf{0.179}\\
	\hline
	%&SPL~\cite{DBLP:conf/ijcai/ZhangZTWCNWY21}      & 0.0036  & 0.0126 & 0.0269 & 0.0143\\
	%\hline
	\multirow{3}{*}{FID $^{\downarrow}$} 
	&PIC~\cite{zheng2019pluralistic}       & 8.82  & 16.44 & 27.82 & 8.71 & 40.42  & 72.43 & 120.10 & 68.00\\
	%&PIC       & 12.28  & 19.45 & 30.53 & 11.76\\
	&DSI~\cite{DBLP:conf/cvpr/Peng0XL21}       & 7.15 & 12.98 & 21.34 & 6.30 & -  & - & - & -\\
	%&SPL~\cite{DBLP:conf/ijcai/ZhangZTWCNWY21}      & 2.71  & 7.71 & 15.81 & 3.71\\
	%&Ours         & \textbf{2.91}  & \textbf{8.21} & \textbf{15.60} & \textbf{3.78}\\
	&Ours         & \textbf{3.22}  & \textbf{8.99} & \textbf{16.54} & \textbf{4.08} & \textbf{34.07}  & \textbf{45.66} & \textbf{78.86} & \textbf{46.97}\\
% 	\hline
    %\cline{2-6}
	%&SPL~\cite{DBLP:conf/ijcai/ZhangZTWCNWY21}      & 2.71  & 7.71 & 15.81 & 3.71\\
	\hline
\end{tabular}
% }
%\vspace{-2mm} DSI~\cite{DBLP:conf/cvpr/Peng0XL21} PIC~\cite{zheng2019pluralistic}
\end{table}

\begin{figure*}[t!]
\begin{center}
\includegraphics[width=1.0\textwidth]{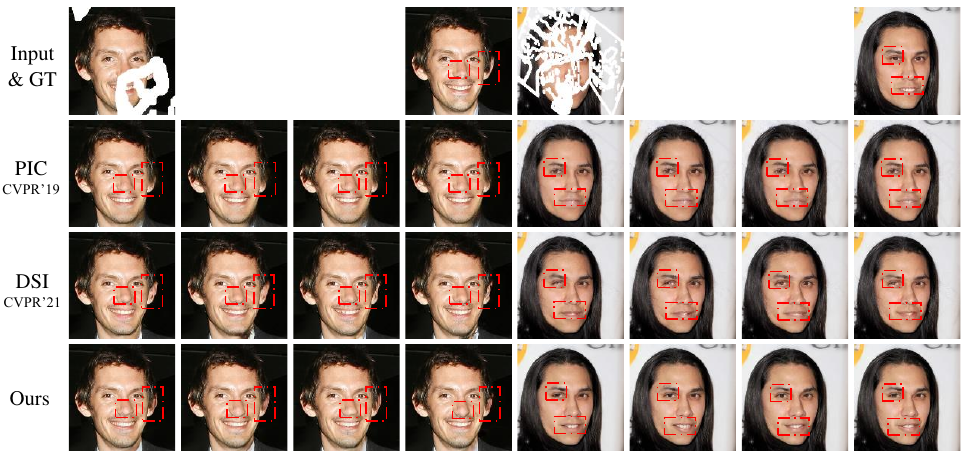}
\end{center}
\vspace{-2mm}
    \caption{Qualitative results of probabilistic image inpainting on the CelebA-HQ dataset. For each input image, we show four generated images by each method, which are obtained with randomly sampled latent variables. 
    }
% \vspace{-1mm}
\label{Qualitative_celebahq}
\end{figure*}

\fig{Qualitative_celebahq} shows four restored images by each compared model given the same input images on the CelebA-HQ dataset.
We can observe that SPN generates more reasonable content than previous approaches. Specifically, images generated by PIC lack diversity among different samples, and results from DSI suffer from distorted structures such as the ears in the upper two rows and the eyes in the bottom two rows.
Instead, human faces restored by SPN not only contain plausible details but also has diverse and reasonable local structures. %, such as the noses in the upper two rows and mouths in the bottom two rows. 
These results show that the stochastic prior pyramid in our model can successfully model the multi-modal distribution of the potential content in the missing area and provide explicit guidance for both global and local restoration.
%
%\textcolor{blue}{Similar observations can also be found on the Paris StreetView dataset in \fig{Qualitative_paris_diverse}.}
%

% \textcolor{red}{TO DO:}

\begin{figure}[t]
\vspace{2mm}
\begin{center}
\includegraphics[width=1.0\linewidth]{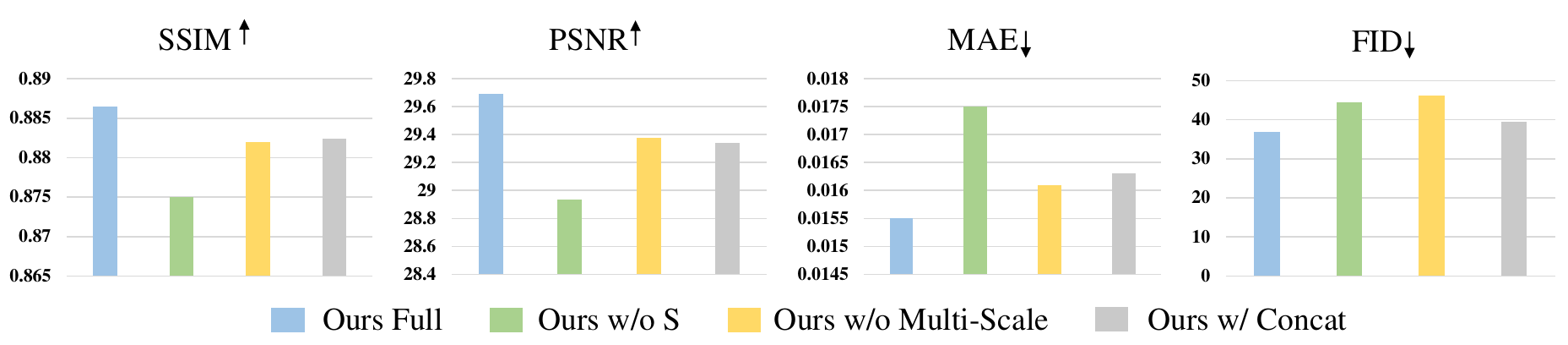}
\end{center}
\vspace{-3mm}
    \caption{Ablation studies on the main components of SPN. Experiments are conducted on the Paris StreetView dataset under the deterministic inpainting setup.} 
% \vspace{-3mm}
\label{img:ab_architeccture}
\end{figure}

\subsection{Ablation Studies}
In this section, we conduct extensive ablation studies to evaluate the contributions proposed in this work. %including 1) the effectiveness of the prior learner based on knowledge distillation and the fully context-aware image generator based on SPADE; and 2) other options of the pretext models to produce the semantic supervisions. 
All experiments are conducted on the Paris StreetView dataset under the deterministic setup. 

\newcommand{\yes}{{\scriptsize\Checkmark}}
\newcommand{\no}{{\scriptsize\XSolidBrush}}

\begin{table}[t]
\caption{\revise{Ablation study for learning semantic priors at different scales. Experiments are conducted on the Paris StreetView dataset under deterministic inpainting setup. The first line denotes our previous work SPL and the last line denotes SPN.}}
\vspace{-5pt}
\footnotesize
\renewcommand{\tabcolsep}{2.0pt}
\centering
\begin{tabular}{|c|c|c|c|c|c|c|}
\hline
64$\times$64 & 128$\times$128 & 256$\times$256 & SSIM$^{\uparrow}$ & PSNR$^{\uparrow}$ & MAE$^{\downarrow}$($\times 10^{-1}$) &FID$^{\downarrow}$\\
\hline
\hline
\yes & \no & \no & 0.882 & 29.38 &0.161 & 46.13 \\
% \hline
\yes & \yes & \no & 0.884 & 29.45 &0.158 & 41.04  \\
% \hline
\yes & \yes & \yes & \textbf{0.886} & \textbf{29.69} & \textbf{0.154} & \textbf{36.82} \\
\hline
\end{tabular}
\label{tab:Ablations_scale}
\vspace{-5pt}
\end{table}

\begin{figure}[t]
\vspace{2mm}
\begin{center}
\includegraphics[width=1.0\linewidth]{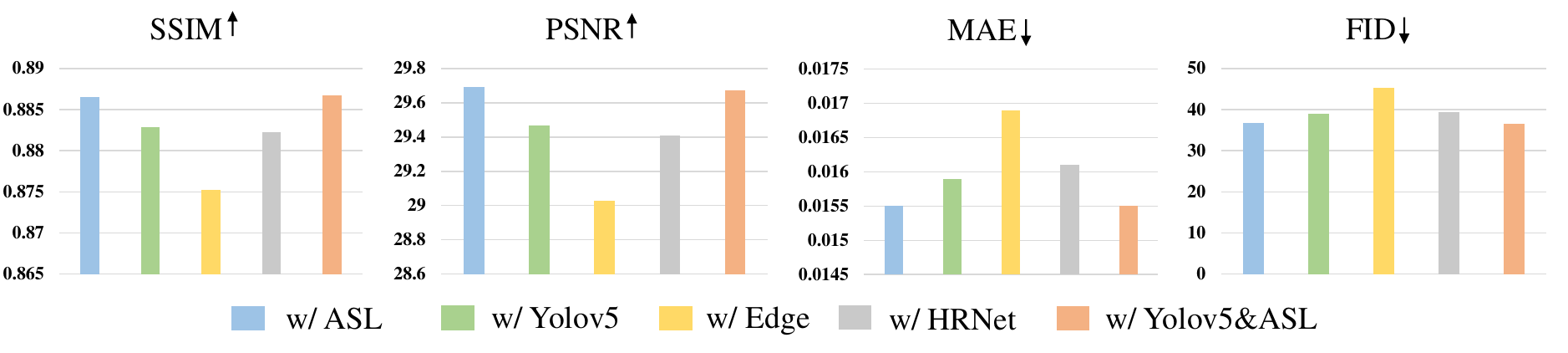}
\end{center}
\vspace{-3mm}
    \caption{Ablation studies on different choices of the semantic supervisions. Experiments are conducted on the Paris StreetView dataset under the deterministic inpainting setup.} 
% \vspace{-3mm}
\label{img:ab_supervision}
\end{figure}

\vspace{5pt}
\subsubsection{On each network component} 

We independently remove different modules of SPN and provide results in \fig{img:ab_architeccture}.
First, we remove all semantic supervisions of knowledge distillation from the prior learner, termed as ``w/o ASL'' (the green bars). By comparing with the blue bars, we observe a significant decline in the performance of SPN. It demonstrates that the pretext knowledge plays an important role in the context-aware image inpainting, and has been effectively transferred to inpainting task.

Besides, we evaluate the effect of multi-scale features in the prior pyramid by only integrating $\boldsymbol{S}^{L}_\text{prior}$ with the smallest spatial size ($64\times 64$) into the image generator, termed as ``w/o Prior Pyramid'' (the yellow bars). In \fig{img:ab_architeccture}, we can observe that replacing the multi-scale priors with single-scale priors leads to a large performance decline.
\revise{In Table~\ref{tab:Ablations_scale}, we further provide a detailed ablation on the usage of semantic priors at different scales. We can observe that as we learn the model with more semantic priors, the FID score is significantly improved. These results show that using multi-scale semantic pyramid indeed help inpainting model generate more perceptually realistic results.}

Finally, we show the effectiveness of the SPADE ResBlocks by simply concatenating image features and the corresponding prior features in the image generator. For example, we have $\boldsymbol{F}^{L}_\text{dec}=\texttt{concat}(\boldsymbol{F}^{L}_\text{enc},\boldsymbol{S}^{L}_\text{prior})$ in Eq. \eqref{eq:ResSPADE}. 
The results are shown by the grey bars and marked as ``w/o SPADE''. We may conclude that the use of SPADE ResBlocks effectively facilitates the joint modeling of both vision features and semantic representations.
\revise{We also conduct ablations on using different numbers of SPADE ResBlocks in Table~\ref{tab:number of blocks}. We can observe that using too less blocks or too many blocks all degenerate the model performance.}

\begin{table}[t]
\vspace{-5pt}
\caption{\revise{Ablation of SPADE ResBlock numbers. Results are obtained on the Paris StreetView dataset. The middle line denotes SPN.}
}
%\vspace{-3mm}
\label{tab:number of blocks}
\footnotesize
\renewcommand{\tabcolsep}{2.0pt}
\centering
\begin{tabular}{l|ccccc}
	%\hline
	%\multicolumn{2}{c|}{Dataset} &\multicolumn{4}{c}{CelebA-HQ}\\
	\hline
	Block number & Params. & SSIM$^{\uparrow}$ & PSNR$^{\uparrow}$ & MAE$^{\downarrow}$($\times 10^{-1}$) &FID$^{\downarrow}$ \\
	\hline 
	4     & 38M  & 0.881 & 29.40 &0.162 & 41.00\\
    8      & 50M & 0.886  & 29.69 &0.154 & 36.82\\
    12      & 62M  & 0.886 & 29.61 &0.155 & 42.50\\
    \hline
\end{tabular}
\vspace{-5pt}
\end{table}

% \vspace{5pt}
\subsubsection{On the choices of semantic supervisions.} 
Since other types of structural priors have also been used by existing inpainting methods, we here explore the alternatives of the semantic supervisions %to learn the prior pyramid
, and see whether they can also facilitate context understanding. 
Specifically, as shown in \fig{img:ab_supervision}, we first replace the multi-scale features from the ASL model, whose results are represented by the blue bars, with those from a detection network YoloV5~\cite{web_reference} pre-trained on the MS COCO dataset~\cite{DBLP:conf/eccv/LinMBHPRDZ14} (the green bars).
Similarly, we also take supervisions from a semantic segmentation model HRNet~\cite{DBLP:journals/pami/00010CJDZ0MTW0X21} pre-trained on the ADE20K dataset~\cite{DBLP:conf/cvpr/ZhouZPFB017} (the grey bars), as well as the edge maps used in the EC method (the yellow bars).
\revise{Notably, these results cannot be strictly compared with each other due to different pre-training datasets and model architectures. However, we can still obtain some interesting observations.
First, the model that uses edge maps as supervisions achieves the worst performance compared with other models using different pre-trained feature maps. Since the edge maps only contain low-level sparse structures instead of high-level semantic information, this result demonstrates that the semantic modeling process can provide more effective guidance for context understanding.
%
% Second, using multi-label classification model or detection model to provide supervisions can produce better results than using segmentation model. Although the pre-training datasets or model architectures are different among these pre-trained models, these results suggest that object-level classification tasks may result in more discriminative representations for scene understanding. We leave this problem in our future work.
Second, object-level tasks such as detection and multi-label classification may result in more discriminative representations for scene understanding.}

\begin{figure}[t]
\begin{center}
\includegraphics[width=1.0\linewidth]{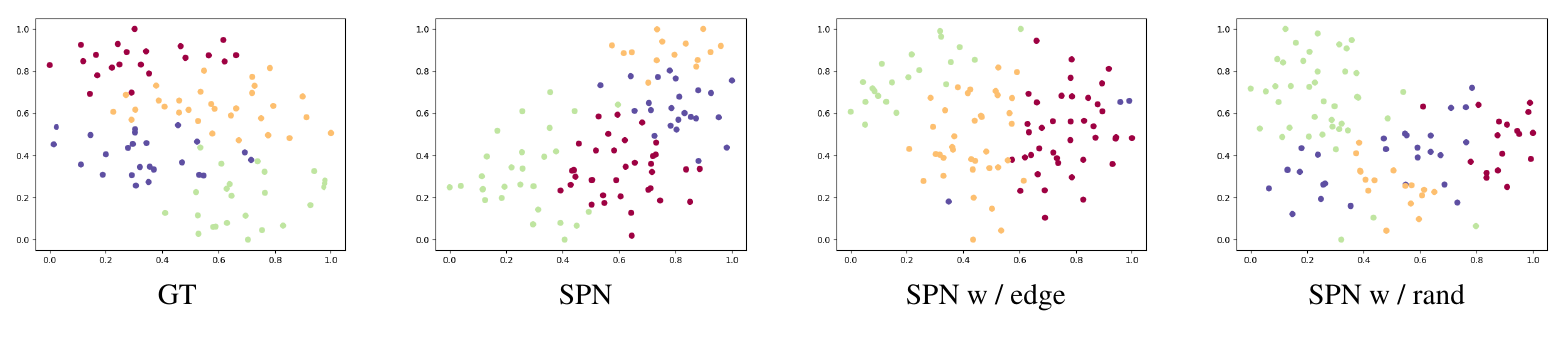}
\end{center}
\vspace{-20pt}
    \caption{\revise{The t-SNE results of different supervisions. Every point denotes a image.}
    }
% \vspace{-1mm}
\label{img:tsne}
\vspace{-5pt}
\end{figure}

\revise{To further evaluate the effectiveness of learning semantic priors from pre-learned networks, we use t-SNE algorithm to visualize the semantic concepts of the generated results. We replace the originally used supervisions with two different features: edge maps and feature maps extracted by a randomly initialized classification model, named ``SPN w / edge'' and ``SPN w / rand'' In Fig~\ref{img:tsne}, we can observe that results from SPN have similar cluster patterns with the real data. On the contrary, results from other ablations may suffer from semantic ambiguities. These results show that with the help of semantic modeling, our approach can generate images with more clear semantic concepts.}

%
% The results demonstrate the superiority of using the multi-label classification model to provide supervisions for the prior pyramid, indicating the significance of transferring object-level scene understanding.
% %
% In contrast, the detection model mainly focuses on particular foreground objects and has a strong bias on the object positions; the edge maps and the segmentation models focus more on object contours instead of object-level semantics.
% % \textcolor{red}{TO DO:}

% Further, we consider to jointly use the detection model (YoloV5) and multi-label classification model (ASL) to provide prior supervisions. It is realized by exploiting two separate sets of distilling heads in the semantic prior learner. 
% %
% We can see that combining these two types of supervisions yields performance comparable to that achieved using the multi-label classification model alone. 

\begin{table}
\caption{Quantitative results on the Paris StreetView dataset using different prior targets. ``w/ $\boldsymbol{S}_\text{prior}^{L}$'' indicates using the $\boldsymbol{S}_\text{prior}^{L}$ feature from another SPN image inpainting model pre-trained on the Places2 dataset. Experiments are conducted under the deterministic setup.
}
\vspace{-1mm}
\label{tab:semantic_information_inSPN}
% \small
\footnotesize
\centering
% \resizebox{\columnwidth}{!}{
\begin{tabular}{l|cccc}
	%\hline
	%\multicolumn{2}{c|}{Dataset} &\multicolumn{4}{c}{Paris StreetView}\\
	%Dataset &\multicolumn{4}{c}{Paris StreetView}\\
	\hline
	%\multicolumn{2}{c|}{Mask Ratio} & 0\%-20\% & 20\%-40\% & 40\%-60\% &All \\
	Metric & SSIM$^{\uparrow}$ & PSNR$^{\uparrow}$ & \revise{MAE$^{\downarrow}$($\times 10^{-1}$)} &FID$^{\downarrow}$ \\
	\hline
	\hline
% 	\multirow{3}{*}{SSIM$^{\uparrow}$} 
	SPL~\cite{DBLP:conf/ijcai/ZhangZTWCNWY21} w/ ASL     & \textbf{0.886} & \textbf{29.38} & \revise{\textbf{0.154}} & 46.13\\
	SPL~\cite{DBLP:conf/ijcai/ZhangZTWCNWY21} w/ $\boldsymbol{S}_\text{prior}^{L}$     &0.881  & 29.21 & \revise{0.163} & \textbf{42.30}\\
	%SPL~\cite{DBLP:conf/ijcai/ZhangZTWCNWY21} w/ $F^{L}_{dec}$     & 0.875 & 28.92 & 0.0170 & 46.87\\
	\hline
\end{tabular}
% }
% \vspace{-2mm}
\end{table}

\subsection{Analyses of the Learned Prior Pyramid}
%In this part, we conduct in-depth analyses of the learned prior pyramid from two different perspectives.

\subsubsection{Semantic knowledge transfer across datasets via the learned prior pyramid}
Since image inpainting can also be utilized as a self-supervised representation learning method~\cite{pathak2016context}, we thus conduct an extra experiment to answer whether the pre-trained SPN model can also provide meaningful semantic representations for image inpainting. Specifically, we use  SPL~\cite{DBLP:conf/ijcai/ZhangZTWCNWY21}, the preliminary approach of SPN, as the baseline model, and replace the original distillation target with $\boldsymbol{S}_\text{prior}^{L}$ from the pre-trained SPN model on the Places2 dataset. The experiment is conducted on the Paris StreetView dataset. From Table~\ref{tab:semantic_information_inSPN}, we can see that using pre-trained $\boldsymbol{S}_\text{prior}^{L}$ as the prior distillation target achieves comparable results with using ASL~\cite{ben2020asymmetric}.
It even performs better in the FID score. Besides, as $\boldsymbol{S}_\text{prior}^{L}$ is only pre-trained on the Places2 dataset, it also shows that the learned representation have potential generalization abilities.

% \vspace{5pt}
\subsubsection{Visualizations of the learned prior pyramid} 
%We respectively visualize the learned priors under both deterministic and probabilistic setups, and the image encoding features before and after being combined with the visual priors. We here extend our previous work with in-depth visualization of the multi-scale representations in the prior pyramid. %Without loss of generality, experiments are conducted under the deterministic inpainting setup.

\begin{figure*}[t]
\begin{center}
\includegraphics[width=1.0\linewidth]{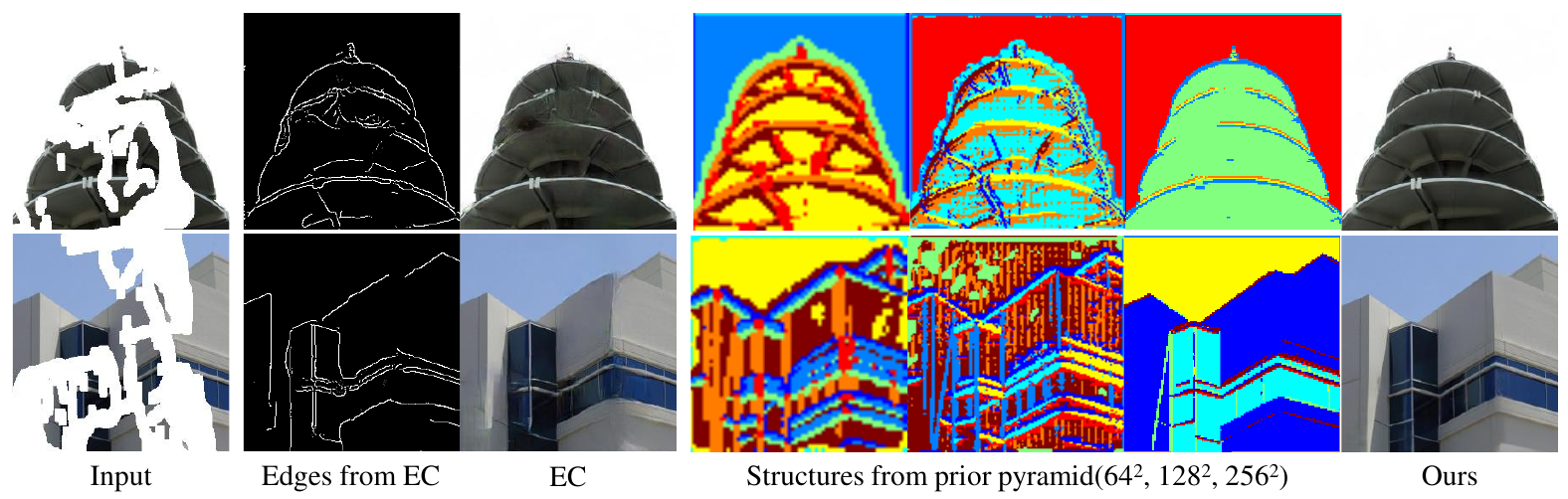}
\end{center}
\vspace{-3mm}
    \caption{Comparisons of the visual priors. We respectively show the edge maps in EC~\protect\cite{nazeri2019edgeconnect} and what has been learned by the semantic prior learner in SPN. We visualize the features in the prior pyramid at different scales (from left to right: $64\times64$, $128\times128$,  $256\times256$) using K-Means clustering. Results are obtained on the Place2 dataset. %Experiments are conducted on the Place2 dataset for deterministic inpainting.
    }
\vspace{-3mm}
\label{img:prior_pyramid}
\end{figure*}

\fig{img:prior_pyramid} compares the structural information generated by the proposed semantic prior learner and the edge maps from the EC model. 
Specifically, we directly perform the K-Means clustering algorithm on the output feature maps and use different colors to represent different clusters. 
\revise{We set $K=8$ for all visualization results and release the clustering function at our Github page.}
From this figure, we observe that the edge maps from the EC model cannot effectively restore reasonable structures for complex scenes, resulting in distorted structures in the final results. 
Instead, the prior learner can capture more complete context understanding and provide consistent structural information at different spatial scales. 
We notice that higher-level semantic priors are more discriminative and are helpful for contextual reasoning, while lower-level priors enhance the understanding of local structure and texture details.

\begin{figure*}[t]
\begin{center}
\includegraphics[width=1.0\linewidth]{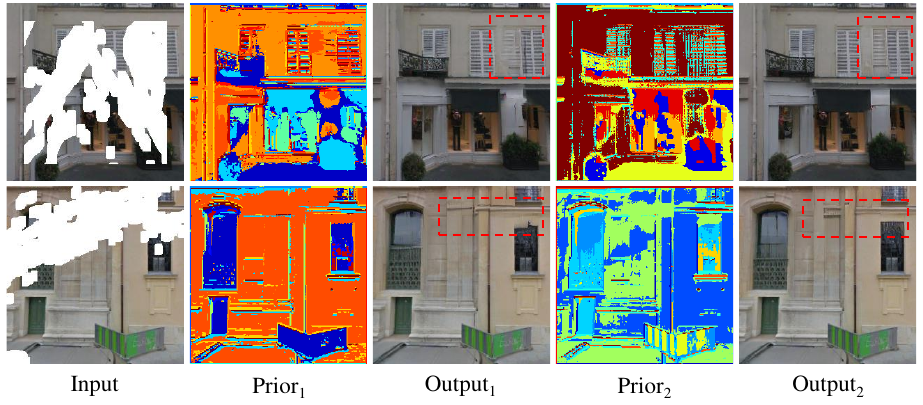}
\end{center}
\vspace{-4mm}
    \caption{Visualizations of randomly sampled priors and corresponding results. ``Prior'' indicates the $\boldsymbol{S}_\text{prior}^{1}$ feature with a spatial size of $256 \times 256$. All showcases are sampled from the Paris StreetView dataset under probabilistic setup. %Zoom in for more details.
    }
\vspace{-3mm}
\label{img:diverse_prior}
\end{figure*}

Besides, in \fig{img:diverse_prior}, we visualize randomly sampled priors in our variational inference module. Given the same input image, we visualize two priors sampled from different latent variables with the largest spatial size ($256 \times 256$). %We also show the corresponding generated images for reference. 
We can observe that the sampled priors contain different structural layout, resulting in diverse details in final output images. These results demonstrate the effectiveness of our variational inference module in handling probabilistic inpainting.

\begin{figure}[t]
\begin{center}
\includegraphics[width=1.0\linewidth]{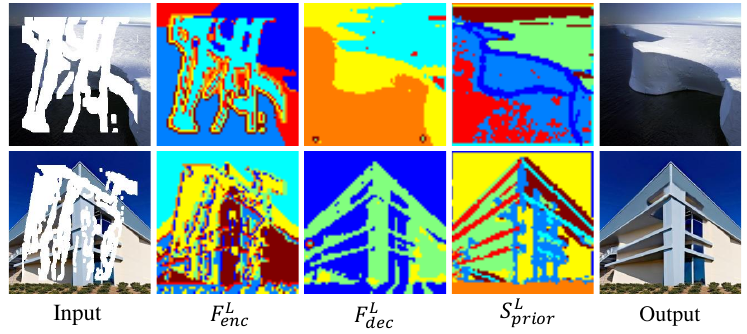}
\end{center}
\vspace{-4mm}
    \caption{The visualization of the feature maps $\boldsymbol{F}^{L}_\text{enc}$ and $\boldsymbol{F}^{L}_\text{dec}$, \textit{before} and \textit{after} the SPADE ResBlocks in the fully context-aware image generator. It reveals the process of feature refinement guided by the high-level semantic priors $\boldsymbol{S}_\text{prior}^{L}$. All showcases are sampled from the Place2 test set.
    }
% \vspace{-3mm}
\label{img:prior_completion}
\end{figure}

In \fig{img:prior_completion}, we further visualize the features in the image generator to show how the feature refinement works by integrating the learned semantic priors. 
We here use
$\boldsymbol{F}^{L}_\text{enc}$ and 
$\boldsymbol{F}^{L}_\text{dec}$ at the smallest spatial scale ($64\times 64$), which are extracted before and after using the $8$ SPADE ResBlocks. We also visualize %$S_{prior}^{L}$
$\boldsymbol{S}_\text{prior}^{L}$ for ease of comparison. 
We can see that $\boldsymbol{F}^{L}_\text{enc}$ is shown to focus more on local texture consistency, while $\boldsymbol{F}^{L}_\text{dec}$ is effectively refined by the learned semantic priors $\boldsymbol{S}_\text{prior}^{L}$ to capture both global context and local texture. 

% \begin{comment}

\subsection{\revise{Analyses of Model Parameters and Inference Time}}

\begin{table*}[t]
\caption{\revise{Comparisons of parameters and inference time. %\textbf{Left}: models for deterministic setup. \textbf{Right}: models for probabilistic setup.
}}
\footnotesize
\label{tab:pamras_time}
\centering
\renewcommand{\tabcolsep}{2.0pt}
\begin{tabular}{l|cccccccc}
	%\hline
	%\multicolumn{2}{c|}{Dataset} &\multicolumn{4}{c}{CelebA-HQ}\\
	\hline
	Model & EC~\cite{nazeri2019edgeconnect}  & MFE~\cite{DBLP:conf/eccv/LiuJSHY20} &RFR~\cite{li2020recurrent} &RN~\cite{yu2020region} &WF~\cite{yu2021wavefill} &CTSDG~\cite{guo2021image} &SPL~\cite{DBLP:conf/ijcai/ZhangZTWCNWY21} &Ours \\%&PIC~\cite{yu2021wavefill} &DSI~\cite{guo2021image} &Ours\\
	\hline
    Params. & 21M & 130M & 30M & 11M & 49M & 52M & 45M & 50M \\
    Time & 0.03s & 0.05s & 0.02s & 0.04s & 0.06s & 0.05s & 0.04s & 0.05s \\
	\hline
\end{tabular}
\end{table*}

%RFR~\cite{li2020recurrent}, RN~\cite{yu2020region}, MFE~\cite{DBLP:conf/eccv/LiuJSHY20}, EC~\cite{nazeri2019edgeconnect}, \revise{WF~\cite{yu2021wavefill}, and CTSDG~\cite{guo2021image}}.DSI~\cite{DBLP:conf/cvpr/Peng0XL21} PIC~\cite{zheng2019pluralistic} 

\revise{The model parameters and single image inference time of different approaches under deterministic setup are shown in Table~\ref{tab:pamras_time}. The inference time is measured on a single GTX3090GPU. On these two metrics, we can observe that our model is comparable with recent approaches such as WF~\cite{yu2021wavefill} and CTSDG~\cite{guo2021image}, and the additional parameters compared with SPL are limited. For probabilistic models, the parameters of PIC~\cite{zheng2019pluralistic}, DSI~\cite{DBLP:conf/cvpr/Peng0XL21}, and ours are respectively $10$M, $256$M, and $68$M, and the inference time are respectively $0.03$s, $25$s, and $0.07$s. The memory and time consumption of our approach are still at a low level.}

\begin{figure}[ht!]
\begin{center}
\includegraphics[width=1.0\linewidth]{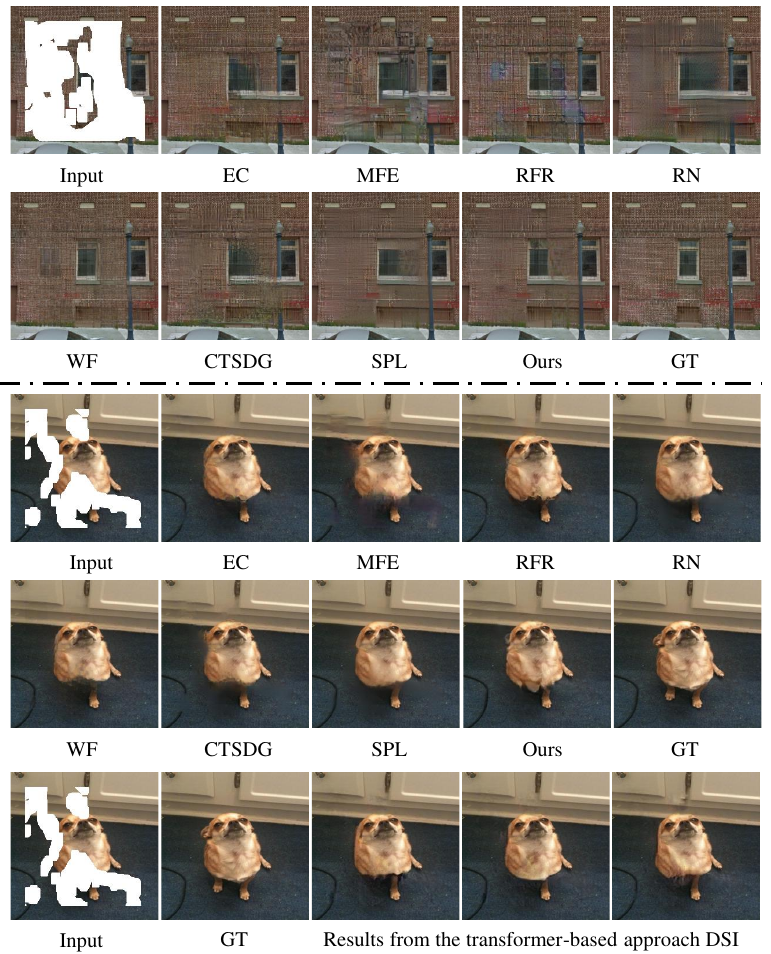}
\end{center}
\vspace{-2mm}
    \caption{\revise{Failure cases on the test sets of Paris StreetView and Places2 respectively.}
    }
\vspace{-1mm}
\label{img:failure}
\end{figure}

\subsection{Limitations}
% In previous sections, we have verified the effectiveness of our method on different datasets.
% %
% However, when large areas of structured information are missing, the generated results are still unsatisfactory.
% %
% In the first case in \fig{img:failure}, we can see that when most of the wall is masked, the generated window in the middle of the output image lacks a clear boundary.
% %
% The problem is even more severe for the RFR~\cite{li2020recurrent} approach.
% %
% For the second case, although our model can generate the overall layout of the dog, one leg is missing due to the lack of a more comprehensive body part understanding.
% %
% One possible direction to solve these problems in the future is to further improve the semantic understanding in image inpainting.
% % \end{comment}

\revise{Although we have verified the effectiveness of our approach on different datasets, there are still some limitations as shown in \fig{img:failure}.
In the first case, we can see that all approaches fail to generate clear window boundaries when large areas of structure information are lost. One possible solution may be using extra user guidance such as sketches to provide explicit constraints.
For the second case, all approaches fail to generate the missing leg, even for the Transformer-based approach DSI~\cite{DBLP:conf/cvpr/Peng0XL21}. We think this case is a typical and also important evidence for the conclusion that existing approaches may not have a complete common sense that dogs usually have four legs, hence these approaches fail to reason the missing leg when it is fully dropped. In addition to introducing extra user guidance, another solution may be using general part-level parser such as Segment Anything~\cite{kirillov2023segment} to provide explicit guidance to help the model be aware of the common sense of different objects.}

\section{Conclusion}
\label{conclusion}

In this paper, we proposed a novel framework named SPN to learn multi-scale semantic priors to help image restoration. Instead of only considering the local texture consistency, we demonstrated that the feature maps in particular pretext models (\textit{e.g.}, multi-label classification and object detection) contain rich semantic information and can help the global context understanding in image inpainting. 
To this end, we proposed a differentiable prior learner to adaptively transfer multi-scale semantic priors through knowledge distillation and extended it to model the multi-modal distributions of missing contents in a probabilistic manner. We also presented a fully context-aware image generator that gradually incorporates the multi-scale prior representations to adaptively refine the image encoding features. %The learned prior pyramid not only provides meaningful hints for global context understanding but also facilitates the local texture restoration in a semantic consistent way. 
In the future work, we aim to introduce the explicit part-level segmentation to help inpainting model be aware of more comprehensive common sense for various objects in real word.

%
% Besides, it can be easily extended to model the multi-modal distributions of missing contents in a probabilistic manner. 
%
% Finally, we presented a fully context-aware image generator that gradually incorporates the multi-scale prior representations to adaptively refine the image encoding features. 
%
% Extensive experiments under both deterministic and probabilistic  setups demonstrated the superiority of SPN on various datasets.

\section*{Acknowledgments} 
This work was supported by NSFC (U19B2035, 62106144, U20B2072, 61976137), Shanghai Municipal Science and Technology Major Project (2021SHZDZX0102), and Shanghai Sailing Program (21Z510202133) from the Science and Technology Commission of Shanghai Municipality.

\bibliography{SPL}

\end{document}